\documentclass[twoside,11pt]{article}

\usepackage{jmlr2e}
\usepackage{mathtools}
\usepackage{color}
\usepackage[caption=false]{subfig}
\usepackage{multirow}
\usepackage{amsmath}
\usepackage{calrsfs}
\usepackage{graphicx}
\usepackage{subfig}

\usepackage[normalem]{ulem}
\useunder{\uline}{\ul}{}

\usepackage{algorithm}
\usepackage{algpseudocode}

\usepackage{array}
\newcolumntype{L}[1]{>{\raggedright\let\newline\\\arraybackslash\hspace{0pt}}m{#1}}
\newcolumntype{C}[1]{>{\centering\let\newline\\\arraybackslash\hspace{0pt}}m{#1}}
\newcolumntype{R}[1]{>{\raggedleft\let\newline\\\arraybackslash\hspace{0pt}}m{#1}}

\DeclareMathAlphabet{\pazocal}{OMS}{zplm}{m}{n}

\ShortHeadings{On the Origin of Deep Learning}{Wang and Raj}
\firstpageno{1}

\begin{document}

\title{On the Origin of Deep Learning}

\author{\name Haohan Wang \email haohanw@cs.cmu.edu 
       \AND
       \name Bhiksha Raj \email bhiksha@cs.cmu.edu \\
       \addr Language Technologies Institute\\
       School of Computer Science\\
       Carnegie Mellon University\\
       }

\editor{---}

\maketitle

\begin{abstract}

This paper is a review of the evolutionary history of deep learning models. It covers from the genesis of neural networks when associationism modeling of the brain is studied, to the models that dominate the last decade of research in deep learning like convolutional neural networks, deep belief networks, and recurrent neural networks. 
In addition to a review of these models, this paper primarily focuses on the precedents of the models above, examining how the initial ideas are assembled to construct the early models and how these preliminary models are developed into their current forms. 
Many of these evolutionary paths last more than half a century and have a diversity of directions. For example, CNN is built on prior knowledge of biological vision system; DBN is evolved from a trade-off of modeling power and computation complexity of graphical models and many nowadays models are neural counterparts of ancient linear models. 
This paper reviews these evolutionary paths and offers a concise thought flow of how these models are developed, and aims to provide a thorough background for deep learning. 
More importantly, along with the path, this paper summarizes the gist behind these milestones and proposes many directions to guide the future research of deep learning.
\end{abstract}


\newpage
\section{Introduction}

Deep learning has dramatically improved the state-of-the-art in many different artificial intelligent tasks like object detection, speech recognition, machine translation \citep{lecun2015deep}. Its deep architecture nature grants deep learning the possibility of solving many more complicated AI tasks \citep{bengio2009learning}. As a result, researchers are extending deep learning to a variety of different modern domains and tasks in additional to traditional tasks like object detection, face recognition, or language models, for example, \cite{osako2015complex} uses the recurrent neural network to denoise speech signals, \cite{gupta2015learning} uses stacked autoencoders to discover clustering patterns of gene expressions. \cite{gatys2015neural} uses a neural model to generate images with different styles. \cite{wang2016select} uses deep learning to allow sentiment analysis from multiple modalities simultaneously, etc. 
This period is the era to witness the blooming of deep learning research. 

However, to fundamentally push the deep learning research frontier forward, one needs to thoroughly understand what has been attempted in the history and why current models exist in present forms. This paper summarizes the evolutionary history of several different deep learning models and explains the main ideas behind these models and their relationship to the ancestors. 
To understand the past work is not trivial as deep learning has evolved over a long time of history, as showed in Table~\ref{tab:history}. Therefore, this paper aims to offer the readers a  walk-through of the major milestones of deep learning research. We will cover the milestones as showed in Table~\ref{tab:history}, as well as many additional works. We will split the story into different sections for the clearness of presentation. 

\begin{table}[]
\centering
\caption{Major milestones that will be covered in this paper}
\label{tab:history}
\begin{tabular}{cC{4cm}L{10cm}}
\hline
Year & Contributer & Contribution \\ \hline
300 BC & Aristotle & introduced Associationism, started the history of human's attempt to understand brain. \\ \hline
1873 & Alexander Bain & introduced Neural Groupings as the earliest models of neural network, inspired Hebbian Learning Rule. \\ \hline
1943 & McCulloch \& Pitts & introduced MCP Model, which is considered as the ancestor of Artificial Neural Model. \\ \hline
1949 & Donald Hebb & considered as the father of neural networks, introduced Hebbian Learning Rule, which lays the foundation of modern neural network. \\ \hline
1958 & Frank Rosenblatt & introduced the first perceptron, which highly resembles modern perceptron. \\ \hline
1974 & Paul Werbos & introduced Backpropagation \\ \hline
\multirow{2}{*}{1980} & Teuvo Kohonen & introduced Self Organizing Map \\ \cline{2-3} 
 & Kunihiko Fukushima & introduced Neocogitron, which inspired Convolutional Neural Network \\ \hline
1982 & John Hopfield & introduced Hopfield Network \\ \hline
1985 & Hilton \& Sejnowski & introduced Boltzmann Machine \\ \hline
\multirow{2}{*}{1986} & Paul Smolensky & introduced Harmonium, which is later known as Restricted Boltzmann Machine \\ \cline{2-3} 
 & Michael I. Jordan & defined and introduced Recurrent Neural Network \\ \hline
 1990 & Yann LeCun & introduced LeNet, showed the possibility of deep neural networks in practice \\ \hline
\multirow{2}{*}{1997} & Schuster \& Paliwal & introduced Bidirectional Recurrent Neural Network \\ \cline{2-3} 
 & Hochreiter \& Schmidhuber & introduced LSTM, solved the problem of vanishing gradient in recurrent neural networks \\ \hline
2006 & Geoffrey Hinton & introduced Deep Belief Networks, also introduced layer-wise pretraining technique, opened current deep learning era. \\ \hline
2009 & Salakhutdinov \& Hinton & introduced Deep Boltzmann Machines \\ \hline
2012 & Geoffrey Hinton & introduced Dropout, an efficient way of training neural networks \\ \hline
\end{tabular}
\end{table}

This paper starts the discussion from research on the human brain modeling. Although the success of deep learning nowadays is not necessarily due to its resemblance of the human brain (more due to its deep architecture), the ambition to build a system that simulate brain indeed thrust the initial development of neural networks. Therefore, the next section begins with connectionism and naturally leads to the age when shallow neural network matures. 

With the maturity of neural networks, this paper continues to briefly discuss the necessity of extending shallow neural networks into deeper ones, as well as the promises deep neural networks make and the challenges deep architecture introduces. 

With the establishment of the deep neural network, this paper diverges into three different popular deep learning topics. Specifically, in Section~\ref{sec:dbn}, this paper elaborates how Deep Belief Nets and its construction component Restricted Boltzmann Machine evolve as a trade-off of modeling power and computation loads. In Section~\ref{sec:cnn}, this paper focuses on the development history of Convolutional Neural Network, featured with the prominent steps along the ladder of ImageNet competition.  In Section~\ref{sec:rnn}, this paper discusses the development of Recurrent Neural Networks, its successors like LSTM, attention models and the successes they achieved. 


While this paper primarily discusses deep learning models, optimization of deep architecture is an inevitable topic in this society. Section~\ref{sec:opt} is devoted to a brief summary of optimization techniques, including advanced gradient method, Dropout, Batch Normalization, etc. 

This paper could be read as a complementary of \citep{schmidhuber2015deep}. Schmidhuber's paper is aimed to assign credit to all those who contributed to the present state of the art, so his paper focuses on every single incremental work along the path, therefore cannot elaborate well enough on each of them. On the other hand, our paper is aimed at providing the background for readers to understand how these models are developed. Therefore, we emphasize on the milestones and elaborate those ideas to help build associations between these ideas. In addition to the paths of classical deep learning models in \citep{schmidhuber2015deep}, we also discuss those recent deep learning work that builds from classical linear models. Another article that readers could read as a complementary is \citep{anderson2000talking} where the authors conducted extensive interviews with well-known scientific leaders in the 90s on the topic of the neural networks' history.

\newpage
\section{From Aristotle to Modern Artificial Neural Networks}
\label{sec:nn}

The study of deep learning and artificial neural networks originates from our ambition to build a computer system simulating the human brain. To build such a system requires understandings of the functionality of our cognitive system. Therefore, this paper traces all the way back to the origins of attempts to understand the brain and starts the discussion of Aristotle's Associationism around 300 B.C. 

\subsection{Associationism}
\begin{quotation}
``When, therefore, we accomplish an act of reminiscence, we pass through a certain series of precursive movements, until we arrive at a movement on which the one we are in quest of is habitually consequent. Hence, too, it is that we hunt through the mental train, excogitating from the present or some other, and from similar or contrary or coadjacent. Through this process reminiscence takes place. For the movements are, in these cases, sometimes at the same time, sometimes parts of the same whole, so that the subsequent movement is already more than half accomplished."

\end{quotation}

This remarkable paragraph of Aristotle is seen as the starting point of Associationism \citep{burnham1888memory}. Associationism is a theory states that mind is a set of conceptual elements that are organized as associations between these elements. Inspired by Plato, Aristotle examined the processes of remembrance and recall and brought up with four laws of association \citep{boeree2000psychology}. 
\begin{itemize}
\item Contiguity: Things or events with spatial or temporal proximity tend to be associated in the mind. 
\item Frequency: The number of occurrences of two events is proportional to the strength of association between these two events. 
\item Similarity: Thought of one event tends to trigger the thought of a similar event. 
\item Contrast: Thought of one event tends to trigger the thought of an opposite event. 
\end{itemize}

Back then, Aristotle described the implementation of these laws in our mind as common sense. For example, the feel, the smell, or the taste of an apple should naturally lead to the concept of an apple, as common sense. Nowadays, it is surprising to see that these laws proposed more than 2000 years ago still serve as the fundamental assumptions of machine learning methods. For example, samples that are near each other (under a defined distance) are clustered into one group; explanatory variables that frequently occur with response variables draw more attention from the model; similar/dissimilar data are usually represented with more similar/dissimilar embeddings in latent space. 

Contemporaneously, similar laws were also proposed by Zeno of Citium, Epicurus and St Augustine of Hippo. The theory of associationism was later strengthened with a variety of philosophers or psychologists. Thomas Hobbes (1588-1679) stated that the complex experiences were the association of simple experiences, which were associations of sensations. He also believed that association exists by means of coherence and frequency as its strength factor. Meanwhile, John Locke (1632-1704) introduced the concept of ``association of ideas''. He still separated the concept of ideas of sensation and ideas of reflection and he stated that complex ideas could be derived from a combination of these two simple ideas. David Hume (1711-1776) later reduced Aristotle's four laws into three: resemblance (similarity), contiguity, and cause and effect. He believed that whatever coherence the world seemed to have was a matter of these three laws. Dugald Stewart (1753-1828) extended these three laws with several other principles, among an obvious one: accidental coincidence in the sounds of words. Thomas Reid (1710-1796) believed that no original quality of mind was required to explain the spontaneous recurrence of thinking, rather than habits. James Mill (1773-1836) emphasized on the law of frequency as the key to learning, which is very similar to later stages of research.  

David Hartley (1705-1757), as a physician, was remarkably regarded as the one that made associationism popular \citep{hartley2013observations}. 
In addition to existing laws, he proposed his argument that memory could be conceived as smaller scale vibrations in the same regions of the brain as the original sensory experience. These vibrations can link up to represent complex ideas and therefore act as a material basis for the stream of consciousness. 
This idea potentially inspired Hebbian learning rule, which will be discussed later in this paper to lay the foundation of neural networks. 

\subsection{Bain and Neural Groupings}

\begin{figure}
\centering
\includegraphics[width=0.5\textwidth]{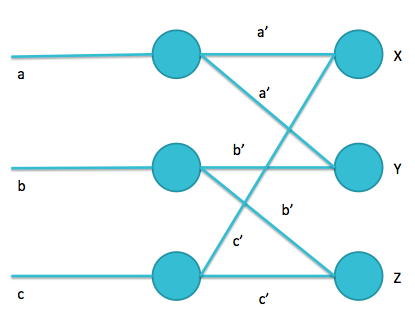}
\caption{Illustration of neural groupings in \citep{bain1873mind}}
\label{fig:bain}
\end{figure}

Besides David Hartley, Alexander Bain (1818-1903) also contributed to the fundamental ideas of Hebbian Learning Rule \citep{wilkes1997bain}. In this book, \cite{bain1873mind} related the processes of associative memory to the distribution of activity of \textit{neural groupings} (a term that he used to denote neural networks back then). He proposed a constructive mode of storage capable of assembling what was required, in contrast to alternative traditional mode of storage with prestored memories. 

To further illustrate his ideas, Bain first described the computational flexibility that allows a neural grouping to function when multiple associations are to be stored. With a few hypothesis, Bain managed to describe a structure that highly resembled the neural networks of today: an individual cell is summarizing the stimulation from other selected linked cells within a grouping, as showed in Figure~\ref{fig:bain}. The joint stimulation from $a$ and $c$ triggers $X$, stimulation from $b$ and $c$ triggers $Y$ and stimulation from $a$ and $c$ triggers $Z$. In his original illustration, $a$, $b$, $c$ stand for simulations, $X$ and $Y$ are outcomes of cells. 

With the establishment of how this associative structure of neural grouping can function as memory, Bain proceeded to describe the construction of these structures. He followed the directions of associationism and stated that relevant impressions of neural groupings must be made in temporal contiguity for a period, either on one occasion or repeated occasions. 

Further, Bain described the computational properties of neural grouping: connections are strengthened or weakened through experience via changes of intervening cell-substance. Therefore, the induction of these circuits would be selected comparatively strong or weak. 

As we will see in the following section, Hebb's postulate highly resembles Bain's description, although nowadays we usually label this postulate as Hebb's, rather than Bain's, according to \citep{wilkes1997bain}. This omission of Bain's contribution may also be due to Bain's lack of confidence in his own theory: Eventually, Bain was not convinced by himself and doubted about the practical values of neural groupings.  

\subsection{Hebbian Learning Rule}
Hebbian Learning Rule is named after Donald O. Hebb (1904-1985) since it was introduced in his work \textit{The Organization of Behavior} \citep{hebb2005organization}. Hebb is also seen as the father of Neural Networks because of this work \citep{didier2011rethinking}. 

In 1949, Hebb stated the famous rule: ``Cells that fire together, wire together", which emphasized on the activation behavior of co-fired cells. More specifically, in his book, he stated that: 
\begin{quotation}
``When an axon of cell A is near enough to excite a cell B and repeatedly or persistently
takes part in firing it, some growth process or metabolic change takes place in one
or both cells such that A’s efficiency, as one of the cells firing B, is increased.''
\end{quotation}
This archaic paragraph can be re-written into modern machine learning languages as the following: 
\begin{align}
\Delta w_i = \eta x_i y
\label{eq:hebb}
\end{align}
where $\Delta w_i$ stands for the change of synaptic weights ($w_i$) of Neuron $i$, of which the input signal is $x_i$. $y$ denotes the postsynaptic response and $\eta$ denotes learning rate. 
In other words, Hebbian Learning Rule states that the connection between two units should be strengthened as the frequency of co-occurrences of these two units increase.   

Although Hebbian Learning Rule is seen as laying the foundation of neural networks, seen today, its drawbacks are obvious: as co-occurrences appear more, the weights of connections keep increasing and the weights of a dominant signal will increase exponentially. This is known as the unstableness of Hebbian Learning Rule \citep{principe1999neural}. Fortunately, these problems did not influence Hebb's identity as the father of neural networks. 

\subsection{Oja's Rule and Principal Component Analyzer}
Erkki Oja extended Hebbian Learning Rule to avoid the unstableness property and he also showed that a neuron, following this updating rule, is approximating the behavior of a Principal Component Analyzer (PCA) \citep{oja1982simplified}. 

Long story short, Oja introduced a normalization term to rescue Hebbian Learning rule, and further he showed that his learning rule is simply an online update of Principal Component Analyzer. We present the details of this argument in the following paragraphs.  

Starting from Equation~\ref{eq:hebb} and following the same notation, Oja showed:
\begin{align*}
w_i^{t+1} = w_i^t + \eta x_i y
\end{align*}
where $t$ denotes the iteration. A straightforward way to avoid the exploding of weights is to apply normalization at the end of each iteration, yielding:
\begin{align*}
w_i^{t+1} = \dfrac{w_i^t + \eta x_i y}{(\sum_{i=1}^n (w_i^t + \eta x_i y)^2)^{\frac{1}{2}}}
\end{align*}
where $n$ denotes the number of neurons. The above equation can be further expanded into the following form:
\begin{align*}
w_i^{t+1} = \dfrac{w_i^t}{Z} + \eta (\dfrac{yx_i}{Z} + \dfrac{w_i\sum_j^nyx_jw_j}{Z^3}) + O(\eta^2)
\end{align*}
where $Z=(\sum_i^nw_i^2)^\frac{1}{2}$. Further, two more assumptions are introduced: 1) $\eta$ is small. Therefore $O(\eta^2)$ is approximately $0$. 2) Weights are normalized, therefore $Z=(\sum_i^nw_i^2)^\frac{1}{2}=1$. 

When these two assumptions were introduced back to the previous equation, Oja's rule was proposed as following:
\begin{align}
w_i^{t+1} = w_i^t + \eta y(x_i-yw_i^t)
\label{eq:oja}
\end{align}

Oja took a step further to show that a neuron that was updated with this rule was effectively performing Principal Component Analysis on the data. 
To show this, Oja first re-wrote Equation~\ref{eq:oja} as the following forms with two additional assumptions \citep{oja1982simplified}:
\begin{align*}
\dfrac{d}{d(t)}w_i^t = Cw_i^t - ((w_i^t)^TCw_i^t)w_i^t
\end{align*}
where $C$ is the covariance matrix of input $X$. Then he proceeded to show this property with many conclusions from his another work \citep{oja1985stochastic} and linked back to PCA with the fact that components from PCA are eigenvectors and the first component is the eigenvector corresponding to largest eigenvalues of the covariance matrix. 
Intuitively, we could interpret this property with a simpler explanation: the eigenvectors of $C$ are the solution when we maximize the rule updating function. Since $w_i^t$ are the eigenvectors of the covariance matrix of $X$, we can get that $w_i^t$ are the PCA. 

Oja's learning rule concludes our story of learning rules of the early-stage neural network. Now we proceed to visit the ideas on neural models. 

\subsection{MCP Neural Model}
While Donald Hebb is seen as the father of neural networks, the first model of neuron could trace back to six years ahead of the publication of Hebbian Learning Rule, when a neurophysiologist Warren McCulloch and a mathematician Walter Pitts speculated the inner workings of neurons and modeled a primitive neural network by electrical circuits based on their findings \citep{mcculloch1943logical}. Their model, known as MCP neural model, was a linear step function upon weighted linearly interpolated data that could be described as: 
\begin{align*}
y =\begin{cases}
      1 \textnormal{, } \sum_iw_ix_i \geq \theta \quad \text{AND} \quad z_j=0, \forall j\\
      0 \textnormal{, otherwise}  
    \end{cases}
\end{align*}
where $y$ stands for output, $x_i$ stands for input of signals, $w_i$ stands for the corresponding weights and $z_j$ stands for the inhibitory input. $\theta$ stands for the threshold. The function is designed in a way that the activity of any inhibitory input completely prevents excitation of
the neuron at any time.

Despite the resemblance between MCP Neural Model and modern perceptron, they are still different distinctly in many different aspects:
\begin{itemize}
\item MCP Neural Model is initially built as electrical circuits. Later we will see that the study of neural networks has borrowed many ideas from the field of electrical circuits. 
\item The weights of MCP Neural Model $w_i$ are fixed, in contrast to the adjustable weights in modern perceptron. All the weights must be assigned with manual calculation. 
\item The idea of inhibitory input is quite unconventional even seen today. It might be an idea worth further study in modern deep learning research. 
\end{itemize}

\subsection{Perceptron}
With the success of MCP Neural Model, Frank Rosenblatt further substantialized Hebbian Learning Rule with the introduction of perceptrons \citep{rosenblatt1958perceptron}. 
While theorists like Hebb were focusing on the biological system in the natural environment, Rosenblatt constructed the electronic device named Perceptron that was showed with the ability to learn in accordance with associationism. 

\begin{figure}[ht]
\centering
\subfloat[Illustration of organization of a perceptron in \citep{rosenblatt1958perceptron}]{
  \includegraphics[clip,width=0.45\textwidth]{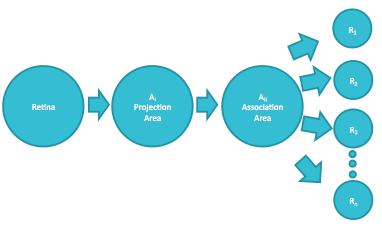}
}
\qquad
\subfloat[A typical perceptron in modern machine learning literature]{
  \includegraphics[clip,width=0.45\textwidth]{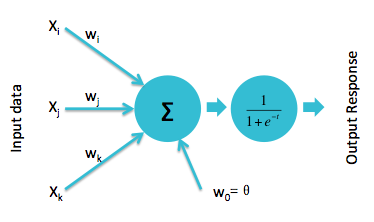}
}
\caption{Perceptrons: (a) A new figure of the illustration of organization of perceptron as in \citep{rosenblatt1958perceptron}. (b) A typical perceptron nowadays, when A$_I$ (Projection Area) is omitted. }
\label{fig:perceptron}
\end{figure}

\cite{rosenblatt1958perceptron} introduced the perceptron with the context of the vision system, as showed in Figure~\ref{fig:perceptron}(a). 
He introduced the rules of the organization of a perceptron as following:
\begin{itemize}
\item Stimuli impact on a retina of the sensory units, which respond in a manner that the pulse amplitude or frequency is proportional to the stimulus intensity. 
\item Impulses are transmitted to \textit{Projection Area} (A$_I$). This projection area is optional. 
\item Impulses are then transmitted to \textit{Association Area} through random connections. If the sum of impulse intensities is equal to or greater than the threshold ($\theta$) of this unit, then this unit fires. 
\item Response units work in the same fashion as those intermediate units. 
\end{itemize}

Figure~\ref{fig:perceptron}(a) illustrates his explanation of perceptron. From left to right, the four units are sensory unit, projection unit, association unit and response unit respectively. Projection unit receives the information from sensory unit and passes onto association unit. This unit is often omitted in other description of similar models. With the omission of projection unit, the structure resembles the structure of nowadays perceptron in a neural network (as showed in Figure~\ref{fig:perceptron}(b)): sensory units collect data, association units linearly adds these data with different weights and apply non-linear transform onto the thresholded sum, then pass the results to response units. 

One distinction between the early stage neuron models and modern perceptrons is the introduction of non-linear activation functions (we use sigmoid function as an example in Figure~\ref{fig:perceptron}(b)). This originates from the argument that linear threshold function should be softened to simulate biological neural networks \citep{bose1996neural} as well as from the consideration of the feasibility of computation to replace step function with a continuous one \citep{mitchell1997machine}. 

After Rosenblatt's introduction of Perceptron, \cite{widrow1960adaptive} introduced a follow-up model called ADALINE. However, the difference between Rosenblatt's Perceptron and ADALINE is mainly on the algorithm aspect. As the primary focus of this paper is neural network models, we skip the discussion of ADALINE. 

\subsection{Perceptron's Linear Representation Power}

A perceptron is fundamentally a linear function of input signals; therefore it is limited to represent linear decision boundaries like the logical operations like NOT, AND or OR, but not XOR when a more sophisticated decision boundary is required. This limitation was highlighted by \cite{minski1969perceptrons}, when they attacked the limitations of perceptions by emphasizing that perceptrons cannot solve functions like XOR or NXOR. As a result, very little research was done in this area until about the 1980’s. 

\begin{figure}
\subfloat[]{\includegraphics[width=0.22\textwidth]{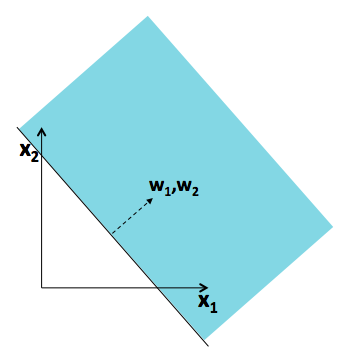}}
\hfill
\subfloat[]{\includegraphics[width=0.22\textwidth]{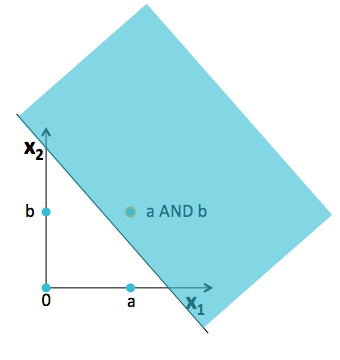}}
\hfill
\subfloat[]{\includegraphics[width=0.22\textwidth]{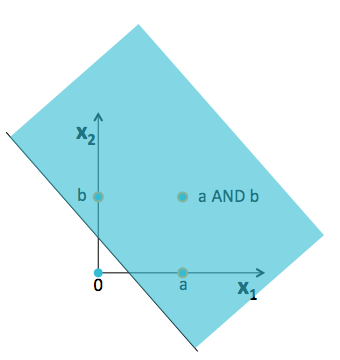}}
\hfill
\subfloat[]{\includegraphics[width=0.22\textwidth]{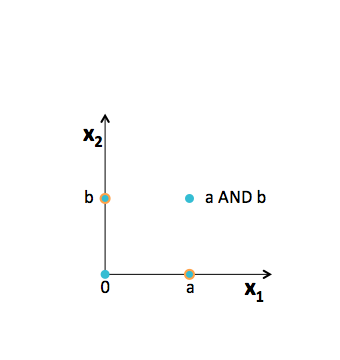}}
\caption{The linear representation power of preceptron}
\label{fig:prep}
\end{figure}

To show a more concrete example, we introduce a linear preceptron with only two inputs $x_1$ and $x_2$, therefore, the decision boundary $w_1x_1 + w_2x_2$ forms a line in a two-dimensional space. The choice of threshold magnitude shifts the line horizontally and the sign of the function picks one side of the line as the halfspace the function represents. The halfspace is showed in Figure~\ref{fig:prep} (a). 

In Figure~\ref{fig:prep} (b)-(d), we present two nodes $a$ and $b$ to denote to input, as well as the node to denote the situation when both of them trigger and a node to denote the situation when neither of them triggers. Figure~\ref{fig:prep} (b) and Figure~\ref{fig:prep} (c) show clearly that a linear perceptron can be used to describe AND and OR operation of these two inputs. However, in Figure~\ref{fig:prep} (d), when we are interested in XOR operation, the operation can no longer be described by a single linear decision boundary. 

In the next section, we will show that the representation ability is greatly enlarged when we put perceptrons together to make a neural network. However, when we keep stacking one neural network upon the other to make a deep learning model, the representation power will not necessarily increase.

\newpage
\section{From Modern Neural Network to the Era of Deep Learning}
\label{deep}
In this section, we will introduce some important properties of neural networks. These properties partially explain the popularity neural network gains these days and also motivate the necessity of exploring deeper architecture. To be specific, we will discuss a set of universal approximation properties, in which each property has its condition. Then, we will show that although a shallow neural network is an universal approximator, deeper architecture can significantly reduce the requirement of resources while retaining the representation power. At last, we will also show some interesting properties discovered in the 1990s about backpropagation, which may inspire some related research today.

\subsection{Universal Approximation Property}
The step from perceptrons to basic neural networks is only placing the perceptrons together. By placing the perceptrons side by side, we get a single one-layer neural network and by stacking one one-layer neural network upon the other, we get a multi-layer neural network, which is often known as multi-layer perceptrons (MLP) \citep{kawaguchi2000multithreaded}.  

One remarkable property of neural networks, widely known as universal approximation property, roughly describes that an MLP can represent any functions. Here we discussed this property in three different aspects:
\begin{itemize}
\item Boolean Approximation: an MLP of one hidden layer\footnote{Through this paper, we will follow the most widely accepted naming convention that calls a two-layer neural network as one hidden layer neural network.} can represent any boolean function exactly. 
\item Continuous Approximation: an MLP of one hidden layer can approximate any bounded continuous function with arbitrary accuracy. 
\item Arbitrary Approximation: an MLP of two hidden layers can approximate any function with arbitrary accuracy. 
\end{itemize}
We will discuss these three properties in detail in the following paragraphs. To suit different readers' interest, we will first offer an intuitive explanation of these properties and then offer the proofs. 

\subsubsection{Representation of any Boolean Functions}
This approximation property is very straightforward. In the previous section we have shown that every linear preceptron can perform either AND or OR. According to De Morgan's laws, every propositional formula can be converted into an equivalent Conjunctive Normal Form, which is an OR of multiple AND functions. Therefore, we simply rewrite the target Boolean function into an OR of multiple AND operations. Then we design the network in such a way: the input layer performs all AND operations, and the hidden layer is simply an OR operation. 

The formal proof is not very different from this intuitive explanation, we skip it for simplicity. 

\subsubsection{Approximation of any Bounded Continuous Functions}
\begin{figure}
\subfloat[]{\includegraphics[width=0.22\textwidth]{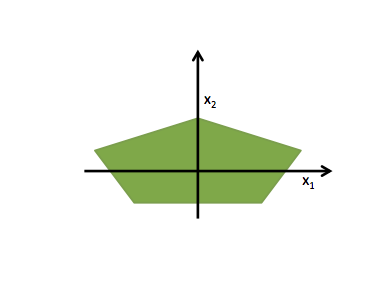}}
\hfill
\subfloat[]{\includegraphics[width=0.22\textwidth]{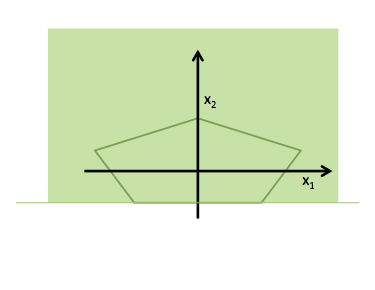}}
\hfill
\subfloat[]{\includegraphics[width=0.22\textwidth]{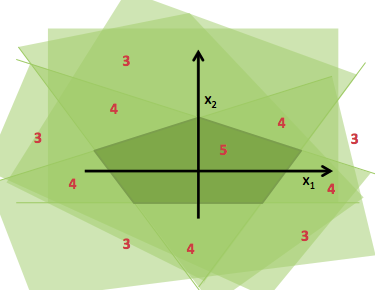}}
\hfill
\subfloat[]{\includegraphics[width=0.22\textwidth]{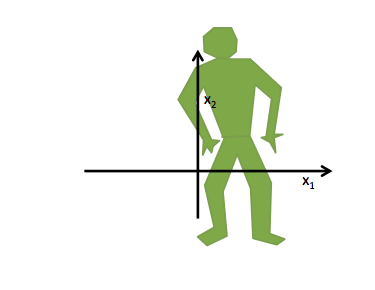}}
\caption{Example of Universal Approximation of any Bounded Continuous Functions}
\label{fig:approx1}
\end{figure}

Continuing from the linear representation power of perceptron discussed previously, if we want to represent a more complex function, showed in Figure~\ref{fig:approx1} (a), we can use a set of linear perceptrons, each of them describing a halfspace. One of these perceptrons is shown in Figure~\ref{fig:approx1} (b), we will need five of these perceptrons. With these perceptrons, we can bound the target function out, as showed in Figure~\ref{fig:approx1} (c). The numbers showed in Figure~\ref{fig:approx1} (c) represent the number of subspaces described by perceptrons that fall into the corresponding region. As we can see, with an appropriate selection of the threshold (e.g. $\theta=5$ in Figure~\ref{fig:approx1} (c)), we can bound the target function out. Therefore, we can describe any bounded continuous function with only one hidden layer; even it is a shape as complicated as Figure~\ref{fig:approx1} (d). 

This property was first shown in \citep{cybenko1989approximation} and \citep{hornik1989multilayer}. To be specific, \cite{cybenko1989approximation} showed that, if we have a function in the following form:
\begin{align}
f(x) = \sum_i \omega_i \sigma (w_i^Tx+\theta)
\label{eq:univ-approximate}
\end{align}
$f(x)$ is dense in the subspace of where it is in. 
In other words, for an arbitrary function $g(x)$ in the same subspace as $f(x)$, we have
\begin{align*}
|f(x)-g(x)| < \epsilon
\end{align*}
where $\epsilon > 0$. 
In Equation~\ref{eq:univ-approximate}, $\sigma$ denotes the activation function (a squashing function back then), $w_i$ denotes the weights for the input layer and $\omega_i$ denotes the weights for the hidden layer. 

This conclusion was drawn with a proof by contradiction: With Hahn-Banach Theorem and Riesz Representation Theorem, the fact that the closure of $f(x)$ is not all the subspace where $f(x)$ is in contradicts the assumption that $\sigma$ is an activation (squashing) function. 

Till today, this property has drawn thousands of citations. Unfortunately, many of the later works cite this property inappropriately \citep{castro2000neural} because Equation~\ref{eq:univ-approximate} is not the widely accepted form of a one-hidden-layer neural network because it does not deliver a thresholded/squashed output, but a linear output instead. Ten years later after this property was shown, \cite{castro2000neural} concluded this story by showing that when the final output is squashed, this universal approximation property still holds. 

Note that, this property was shown with the context that activation functions are squashing functions. By definition, a squashing function $\sigma: R \rightarrow [0, 1]$ is a non-decreasing function with the properties $\lim_{x \rightarrow \infty} \sigma(x) = 1$ and $\lim_{x \rightarrow -\infty} \sigma(x) = 0$. Many activation functions of recent deep learning research do not fall into this category. 

\subsubsection{Approximation of Arbitrary Functions}
\begin{figure}
\centering
\includegraphics[width=0.8\textwidth]{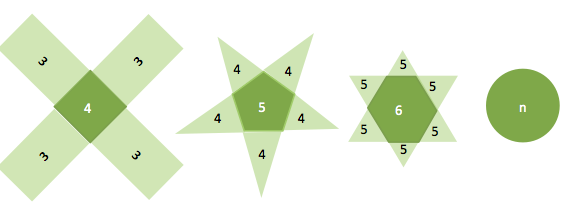}
\caption{Threshold is not necessary with a large number of linear perceptrons.}
\label{fig:ua21}
\end{figure}

Before we move on to explain this property, we need first to show a major property regarding combining linear perceptrons into neural networks. Figure~\ref{fig:ua21} shows that as the number of linear perceptrons increases to bound the target function, the area outside the polygon with the sum close to the threshold shrinks. Following this trend, we can use a large number of perceptrons to bound a circle, and this can be achieved even without knowing the threshold because the area close to the threshold shrinks to nothing. What left outside the circle is, in fact, the area that sums to $\frac{N}{2}$, where $N$ is the number of perceptrons used. 

\begin{figure}
\centering
\includegraphics[width=0.8\textwidth]{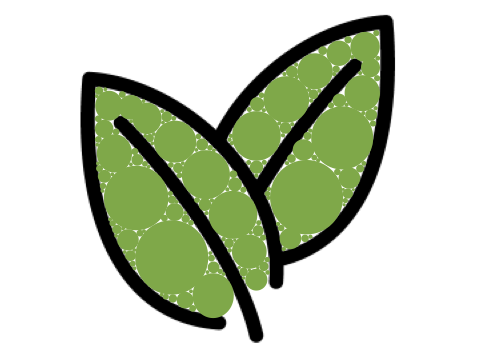}
\caption{How a neural network can be used to approximate a leaf shaped function.}
\label{fig:ua}
\end{figure}

Therefore, a neural network with one hidden layer can represent a circle with arbitrary diameter. Further, we introduce another hidden layer that is used to combine the outputs of many different circles. This newly added hidden layer is only used to perform OR operation. Figure~\ref{fig:ua} shows an example that when the extra hidden layer is used to merge the circles from the previous layer, the neural network can be used to approximate any function. The target function is not necessarily continuous. However, each circle requires a large number of neurons, consequently, the entire function requires even more. 

This property was showed in \citep{lapedes1988neural} and \citep{university1988continuous} respectively. Looking back at this property today, it is not arduous to build the connections between this property to Fourier series approximation, which, in informal words, states that every function curve can be decomposed into the sum of many simpler curves. With this linkage, to show this universal approximation property is to show that any one-hidden-layer neural network can represent one simple surface, then the second hidden layer sums up these simple surfaces to approximate an arbitrary function. 

As we know, one hidden layer neural network simply performs a thresholded sum operation, therefore, the only step left is to show that the first hidden layer can represent a simple surface. To understand the ``simple surface'', with linkage to Fourier transform, one can imagine one cycle of the sinusoid for the one-dimensional case or a ``bump'' of a plane in the two-dimensional case. 

For one dimension, to create a simple surface, we only need two sigmoid functions appropriately placed, for example, as following:
\begin{align*}
f_1(x) &= \dfrac{h}{1+e^{-(x+t_1)}} \\
f_2(x) &= \dfrac{h}{1+e^{x-t_2}}
\end{align*}
Then, with $f_1(x)+f_2(x)$, we create a simple surface with height $2h$ from $t_1 \leq x \leq t_2$. This could be easily generalized to $n$-dimensional case, where we need $2n$ sigmoid functions (neurons) for each simple surface. Then for each simple surface that contributes to the final function, one neuron is added onto the second hidden layer. Therefore, despite the number of neurons need, one will never need a third hidden layer to approximate any function. 

Similarly to how Gibbs phenomenon affects Fourier series approximation, this approximation cannot guarantee an exact representation. 

The universal approximation properties showed a great potential of shallow neural networks at the price of exponentially many neurons at these layers. One followed-up question is that how to reduce the number of required neurons while maintaining the representation power. This question motivates people to proceed to deeper neural networks despite that shallow neural networks already have infinite modeling power. Another issue worth attention is that, although neural networks can approximate any functions, it is not trivial to find the set of parameters to explain the data. In the next two sections, we will discuss these two questions respectively. 

\subsection{The Necessity of Depth}
The universal approximation properties of shallow neural networks come at a price of exponentially many neurons and therefore are not realistic. The question about how to maintain this expressive power of the network while reducing the number of computation units has been asked for years.
Intuitively, \cite{bengio2011expressive} suggested that it is nature to pursue deeper networks because 1) human neural system is a deep architecture (as we will see examples in Section~\ref{sec:cnn} about human visual cortex.) and 2) humans tend to represent concepts at one level of abstraction as the composition of concepts at lower levels. 
Nowadays, the solution is to build deeper architectures, which comes from a conclusion that states the representation power of a $k$ layer neural network with polynomial many neurons need to be expressed with exponentially many neurons if a $k-1$ layer structured is used. However, theoretically, this conclusion is still being completed. 

This conclusion could trace back to three decades ago when \cite{yao1985separating} showed the limitations of shallow circuits functions. \cite{hastad1986almost} later showed this property with parity circuits: ``there are functions computable in polynomial size and depth $k$ but requires exponential size when depth is restricted to $k-1$''. He showed this property mainly by the application of DeMorgan's law, which states that any AND or ORs can be rewritten as OR of ANDs and vice versa. Therefore, he simplified a circuit where ANDs and ORs appear one after the other by rewriting one layer of ANDs into ORs and therefore merge this operation to its neighboring layer of ORs. By repeating this procedure, he was able to represent the same function with fewer layers, but more computations. 

Moving from circuits to neural networks, \cite{delalleau2011shallow} compared deep and shallow sum-product neural networks. They showed that a function that could be expressed with $O(n)$ neurons on a network of depth $k$ required at least $O(2^{\sqrt{n}})$ and $O((n-1)^k)$ neurons on a two-layer neural network. 

Further, \cite{bianchini2014complexity} extended this study to a general neural network with many major activation functions including \textit{tanh} and \textit{sigmoid}. They derived the conclusion with the concept of Betti numbers, and used this number to describe the representation power of neural networks. They showed that for a shallow network, the representation power can only grow polynomially with respect to the number of neurons, but for deep architecture, the representation can grow exponentially with respect to the number of neurons. They also related their conclusion to VC-dimension of neural networks, which is $O(p^2)$ for \textit{tanh} \citep{bartlett2003vapnik} where $p$ is the number of parameters. 

Recently, \cite{eldan2015power} presented a more thorough proof to show that depth of a neural network is exponentially more valuable than the width of a neural network, for a standard MLP with any popular activation functions. Their conclusion is drawn with only a few weak assumptions that constrain the activation functions to be mildly increasing, measurable, and able to allow shallow neural networks to approximate any univariate Lipschitz function. Finally, we have a well-grounded theory to support the fact that deeper network is preferred over shallow ones. However, in reality, many problems will arise if we keep increasing the layers. Among them, the increased difficulty of learning proper parameters is probably the most prominent one. Immediately in the next section, we will discuss the main drive of searching parameters for a neural network: Backpropagation. 

\subsection{Backpropagation and Its Properties}
Before we proceed, we need to clarify that the name backpropagation, originally, is not referring to an algorithm that is used to learn the parameters of a neural network, instead, it stands for a technique that can help efficiently compute the gradient of parameters when gradient descent algorithm is applied to learn parameters \citep{hecht1989theory}. However, nowadays it is widely recognized as the term to refer gradient descent algorithm with such a technique. 

Compared to a standard gradient descent, which updates all the parameters with respect to error, backpropagation first propagates the error term at output layer back to the layer at which parameters need to be updated, then uses standard gradient descent to update parameters with respect to the propagated error. Intuitively, the derivation of backpropagation is about organizing the terms when the gradient is expressed with the chain rule. The derivation is neat but skipped in this paper due to the extensive resources available \citep{werbos1990backpropagation,mitchell1997machine,lecun2015deep}.  
Instead, we will discuss two interesting and seemingly contradictory properties of backpropagation. 

\subsubsection{Backpropagation Finds Global Optimal for Linear Separable Data}
\cite{gori1992problem} studied on the problem of local minima in backpropagation. Interestingly, when the society believes that neural networks or deep learning approaches are believed to suffer from local optimal, they proposed an architecture where global optimal is guaranteed. Only a few weak assumptions of the network are needed to reach global optimal, including
\begin{itemize}
\item Pyramidal Architecture: upper layers have fewer neurons
\item Weight matrices are full row rank
\item The number of input neurons cannot smaller than the classes/patterns of data. 
\end{itemize}
However, their approaches may not be relevant anymore as they require the data to be linearly separable, under which condition that many other models can be applied.  

\subsubsection{Backpropagation Fails for Linear Separable Data}
On the other hand, \cite{brady1989back} studied the situations when backpropagation fails on linearly separable data sets. He showed that there could be situations when the data is linearly separable, but a neural network learned with backpropagation cannot find that boundary. He also showed examples when this situation occurs. 

His illustrative examples only hold when the misclassified data samples are significantly less than correctly classified data samples, in other words, the misclassified data samples might be just outliers. Therefore, this interesting property, when viewed today, is arguably a desirable property of backpropagation as we typically expect a machine learning model to neglect outliers. Therefore, this finding has not attracted many attentions. 

However, no matter whether the data is an outlier or not, neural network should be able to overfit training data given sufficient training iterations and a legitimate learning algorithm, especially considering that \cite{brady1989back} showed that an inferior algorithm was able to overfit the data. Therefore, this phenomenon should have played a critical role in the research of improving the optimization techniques. Recently, the studying of cost surfaces of neural networks have indicated the existence of saddle points \citep{choromanska2015loss,dauphin2014identifying,pascanu2014saddle}, which may explain the findings  of Brady \textit{et al} back in the late 80s. 

Backpropagation enables the optimization of deep neural networks. However, there is still a long way to go before we can optimize it well. Later in Section~\ref{sec:opt}, we will briefly discuss more techniques related to the optimization of neural networks. 
\newpage
\section{The Network as Memory and Deep Belief Nets}
\label{sec:dbn}
\begin{figure}[ht]
\centering
\includegraphics[width=0.8\textwidth]{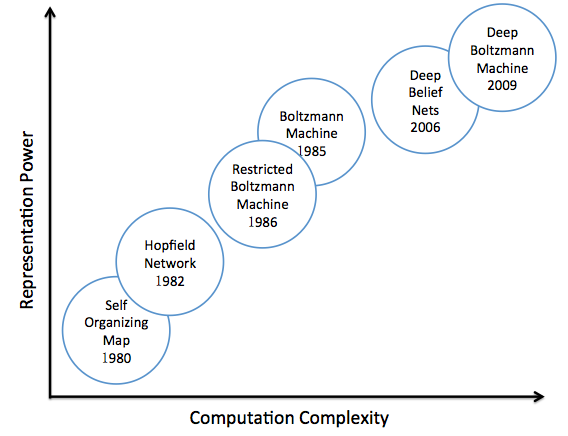}
\caption{Trade off of representation power and computation complexity of several models, that guides the development of better models}
\label{fig:memory}
\end{figure}
With the background of how modern neural network is set up, we proceed to visit the each prominent branch of current deep learning family. Our first stop is the branch that leads to the popular Restricted Boltzmann Machines and Deep Belief Nets, and it starts as a model to understand the data unsupervisedly. 

Figure~\ref{fig:memory} summarizes the model that will be covered in this Section. The horizontal axis stands for the computation complexity of these models while the vertical axis stands for the representation power. The six milestones that will be focused in this section are placed in the figure. 

\subsection{Self Organizing Map}
The discussion starts with Self Organizing Map (SOM) invented by \cite{kohonen1990self}. SOM is a powerful technique that is primarily used in reducing the dimension of data, usually to one or two dimensions \citep{germano1999self}. While reducing the dimensionality, SOM also retains the topological similarity of data points. It can also be seen as a tool for clustering while imposing the topology on clustered representation. 
\begin{figure}
\centering
\includegraphics[width=0.8\textwidth]{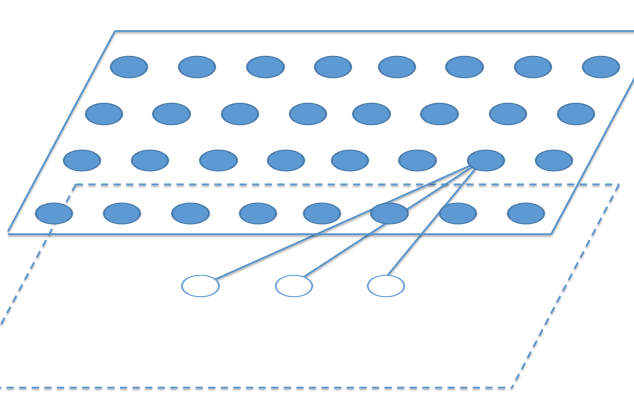}
\caption{Illustration of Self-Organizing Map}
\label{fig:som}
\end{figure}
Figure~\ref{fig:som} is an illustration of Self Organizing Map of two dimension hidden neurons. Therefore, it learns a two dimension representation of data. The upper shaded nodes denote the units of SOM that are used to represent data while the lower circles denote the data. There is no connection between the nodes in SOM
\footnote{In some other literature, \citep{bullinaria2004self} as an example, one may notice that there are connections in the illustrations of models. However, those connections are only used to represent the neighborhood relationship of nodes, and there is no information flowing via those connections. In this paper, as we will show many other models that rely on a clear illustration of information flow, we decide to save the connections to denote that.}. 

The position of each node is fixed. The representation should not be viewed as only a numerical value. Instead, the position of it also matters. This property is different from some widely-accepted representation criterion. For example, we compare the case when one-hot vector and one-dimension SOM are used to denote colors: To denote \textit{green} out of a set: $C=\{green, red, purple\}$, one-hot representation can use any vector of $(1,0,0)$, $(0,1,0)$ or $(0,0,1)$ as long as we specify the bit for \textit{green} correspondingly. However, for a one-dimensional SOM, only two vectors are possible: $(1,0,0)$ or $(0,0,1)$. This is because that, since SOM aims to represent the data while retaining the similarity; and \textit{red} and \textit{purple} are much more similar than \textit{green} and \textit{red} or \textit{green} and \textit{purple}, \textit{green} should not be represented in a way that it splits \textit{red} and \textit{purple}. One should notice that, this example is only used to demonstrate that the position of each unit in SOM matters. In practice, the values of SOM unit are not restricted to integers. 

The learned SOM is usually a good tool for visualizing data. For example, if we conduct a survey on the happiness level and richness level of each country and feed the data into a two-dimensional SOM. Then the trained units should represent the happiest and richest country at one corner and represent the opposite country at the furthest corner. The rest two corners represent the richest, yet unhappiest and the poorest but happiest countries. The rest countries are positioned accordingly. The advantage of SOM is that it allows one to easily tell how a country is ranked among the world with a simple glance of the learned units \citep{guthikonda2005kohonen}.

\subsubsection{Learning Algorithm}
With an understanding of the representation power of SOM, now we proceed to its parameter learning algorithm. The classic algorithm is heuristic and intuitive, as shown below:
\begin{algorithm}
\begin{algorithmic}
\State Initialize weights of all units, $w_{i,j}$ $\forall$ i, j
\For{$t \leq N$} 
\State Pick $v_k$ randomly
\State Select Best Matching Unit (BMU) as $p, q := \arg\min_{i,j}||w_{ij}-v_k||_2^2$
\State Select the nodes of interest as the neighbors of BMU. $I=\{w_{i,j} | \textnormal{dist}(w_{i,j}, w_{p,q}) < r(t) \}$
\State Update weights: $w_{i,j}=w_{i,j}+P(i,j,p,q)l(t)||w_{ij}-v_k||_2^2$, $\forall i,j \in I$
\EndFor
\end{algorithmic}
\end{algorithm}
Here we use a two-dimensional SOM as example, and $i,j$ are indexes of units; $w$ is weight of the unit; $v$ denotes data vector; $k$ is the index of data; $t$ denotes the current iteration; $N$ constrains the maximum number of steps allowed; $P(\cdot)$ denotes the penalty considering the distance between unit $p,q$ and unit $i,j$; $l$ is learning rate; $r$ denotes a radius used to select neighbor nodes. Both $l$ and $r$ typically decrease as $t$ increases. $||\cdot||_2^2$ denotes Euclidean distance and $\textnormal{dist}(\cdot)$ denotes the distance on the position of units. 

This algorithm explains how SOM can be used to learn a representation and how the similarities are retained as it always selects a subset of units that are similar with the data sampled and adjust the weights of units to match the data sampled. 

However, this algorithm relies on a careful selection of the radius of neighbor selection and a good initialization of weights. Otherwise, although the learned weights will have a local property of topological similarity, it loses this property globally: sometimes, two similar clusters of similar events are separated by another dissimilar cluster of similar events. In simpler words, units of \textit{green} may actually separate units of \textit{red} and units of \textit{purple} if the network is not appropriately trained. \citep{germano1999self}. 

\subsection{Hopfield Network}
Hopfield Network is historically described as a form of recurrent\footnote{The term ``recurrent'' is very confusing nowadays because of the popularity recurrent neural network (RNN) gains.} neural network, first introduced in \citep{hopfield1982neural}. ``Recurrent'' in this context refers to the fact that the weights connecting the neurons are bidirectional. Hopfield Network is widely recognized because of its content-addressable memory property. This content-addressable memory property is a simulation of the spin glass theory. Therefore, we start the discussion from spin glass. 

\subsubsection{Spin Glass}
The spin glass is physics term that is used to describe a magnetic phenomenon. Many works have been done for a detailed study of related theory \citep{edwards1975theory, mezard1990spin}, so in this paper, we only describe this it intuitively. 

When a group of dipoles is placed together in any space. Each dipole is forced to align itself with the field generated by these dipoles at its location. However, by aligning itself, it changes the field at other locations, leading other dipoles to flip, causing the field in the original location to change. Eventually, these changes will converge to a stable state. 

To describe the stable state, we first define the total field at location $j$ as
\begin{align*}
s_j = o_j + c^t\sum_k  \dfrac{s_k}{d^2_{jk}}
\end{align*}
where $o_j$ is an external field, $c^t$ is a constant that depends on temperature $t$, $s_k$ is the polarity of the $k$th dipole and $d_{jk}$ is the distance from location $j$ to location $k$. Therefore, the total potential energy of the system is:
\begin{align}
PE =  \sum_j   s_j o_j  +  c^ts_j \sum_k \dfrac{s_k}{d^2_{jk}}
\label{eq:spin}
\end{align}
The magnetic system will evolve until this potential energy is minimum.

\subsubsection{Hopfield Network}

\begin{figure}
\centering
\includegraphics[width=0.8\textwidth]{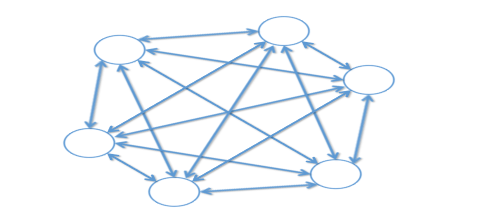}
\caption{Illustration of Hopfield Network. It is a fully connected network of six binary thresholding neural units. Every unit is connected with data, therefore these units are denoted as unshaded nodes. }
\label{fig:hopfield}
\end{figure}

Hopfield Network is a fully connected neural network with binary thresholding neural units. The values of these units are either 0 or 1\footnote{Some other literature may use -1 and 1 to denote the values of these units. While the choice of values does not affect the idea of Hopfiled Network, it changes the formulation of energy function. In this paper, we only discuss in the context of 0 and 1 as values. }. These units are fully connected with bidirectional weights. 

With this setting, the energy of a Hopfield Network is defined as:
\begin{align}
E = -\sum_is_ib_i - \sum_{i,j}s_is_jw_{i,j}
\label{eq:hopfield}
\end{align}
where $s$ is the state of a unit, $b$ denotes the bias; $w$ denotes the bidirectional weights and $i,j$ are indexes of units. This energy function closely connects to the potential energy function of spin glass, as showed in Equation~\ref{eq:spin}. 

Hopfield Network is typically applied to memorize the state of data. The weights of a network are designed or learned to make sure that the energy is minimized given the state of interest. Therefore, when another state presented to the network, while the weights are fixed, Hopfield Network can search for the states that minimize the energy and recover the state in memory. For example, in a face completion task, when some image of faces are presented to Hopfield Network (in a way that each unit of the network corresponds to each pixel of one image, and images are presented one after the other), the network can calculate the weights to minimize the energy given these faces. Later, if one image is corrupted or distorted and presented to this network again, the network is able to recover the original image by searching a configuration of states to minimize the energy starting from corrupted input presented. 

The term ``energy'' may remind people of physics. To explain how Hopfield Network works in a physics scenario will be clearer: nature uses Hopfield Network to memorize the equilibrium position of a pendulum because, in an equilibrium position, the pendulum has the lowest gravitational potential energy. Therefore, whenever a pendulum is placed, it will converge back to the equilibrium position. 

\subsubsection{Learning and Inference}
Learning of the weights of a Hopfield Network is straightforward \citep{gurney1997introduction}. The weights can be calculated as:
\begin{align*}
w_{i,j} = \sum_{i,j}(2s_i-1)(2s_j-1)
\end{align*}
the notations are the same as Equation~\ref{eq:hopfield}. 

This learning procedure is simple, but still worth mentioning as it is an essential step of a Hopfield Network when it is applied to solve practical problems. However, we find that many online tutorials omit this step, and to make it worse, they refer the inference of states as learning/training. To remove the confusion, in this paper, similar to how terms are used in standard machine learning society, we refer the calculation of weights of a model (either from closed-form solution, or numerical solution) as ``parameter learning'' or ``training''. We refer the process of applying an existing model with weights known onto solving a real-world problem as ``inference''\footnote{``inference'' is conventionally used in such a way in machine learning society, although some statisticians may disagree with this usage.} or ``testing'' (to decode a hidden state of data, e.g. to predict a label).

The inference of Hopfield Network is also intuitive. For a state of data, the network tests that if inverting the state of one unit, whether the energy will decrease. If so, the network will invert the state and proceed to test the next unit. This procedure is called \textbf{Asynchronous} update and this procedure is obviously subject to the sequential order of selection of units. A counterpart is known as \textbf{Synchronous} update when the network first tests for all the units and then inverts all the unit-to-invert simultaneously. Both of these methods may lead to a local optimal. Synchronous update may even result in an increasing of energy and may converge to an oscillation or loop of states. 

\subsubsection{Capacity}
One distinct disadvantage of Hopfield Network is that it cannot keep the memory very efficient because a network of $N$ units can only store memory up to $0.15N^2$ bits. While a network with $N$ units has $N^2$ edges. In addition, after storing $M$ memories ($M$ instances of data), each connection has an integer value in range $[-M, M]$. Thus, the number of bits required to store $N$ units are $N^2log(2M+1)$ \citep{hopfield1982neural}. Therefore, we can safely draw the conclusion that although Hopfield Network is a remarkable idea that enables the network to memorize data, it is extremely inefficient in practice. 

As follow-ups of the invention of Hopfield Network, many works are attempted to study and increase the capacity of original Hopfield Network \citep{storkey1997increasing,liou1999error,liou2006finite}. Despite these attempts made, Hopfield Network still gradually fades out of the society. It is replaced by other models that are inspired by it. Immediately following this section, we will discuss the popular Boltzmann Machine and Restricted Boltzmann Machine and study how these models are upgraded from the initial ideas of Hopfield Network and evolve to replace it. 

\subsection{Boltzmann Machine}
Boltzmann Machine, invented by \citet{ackley1985learning}, is a stochastic with-hidden-unit version Hopfield Network. It got its name from Boltzmann Distribution. 

\subsubsection{Boltzmann Distribution}
Boltzmann Distribution is named after Ludwig Boltzmann and investigated extensively by \citep{willard1902elementary}. It is originally used to describe the probability distribution of particles in a system over various possible states as following:
\begin{align*}
F(s) \propto e^{-\frac{E_s}{kT}}
\end{align*}
where $s$ stands for the state and $E_s$ is the corresponding energy. $k$ and $T$ are Boltzmann's constant and thermodynamic temperature respectively. Naturally, the ratio of two distribution is only characterized by the difference of energies, as following: 
\begin{align*}
r = \dfrac{F(s_1)}{F(s_2)} = e^{\frac{E_{s_2} - E_{s_1}}{kT}}
\end{align*}
which is known as \textbf{Boltzmann factor}. 

With how the distribution is specified by the energy, the probability is defined as the term of each state divided by a normalizer, as following: 
\begin{align*}
p_{s_i} = \dfrac{e^{-\frac{E_{s_i}}{kT}}}{\sum_je^{-\frac{E_{s_j}}{kT}}}
\end{align*}

\subsubsection{Boltzmann Machine}
\begin{figure}
\centering
\includegraphics[width=0.8\textwidth]{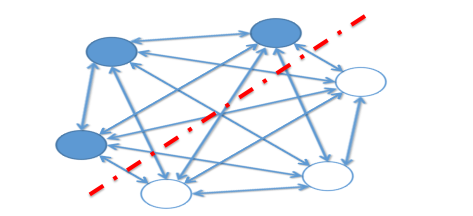}
\caption{Illustration of Boltzmann Machine. With the introduction of hidden units (shaded nodes), the model conceptually splits into two parts: visible units and hidden units. The red dashed line is used to highlight the conceptual separation. }
\label{fig:bm}
\end{figure}

As we mentioned previously, Boltzmann Machine is a stochastic with-hidden-unit version Hopfield Network. Figure~\ref{fig:bm} introduces how the idea of hidden units is introduced that turns a Hopfield Network into a Boltzmann Machine. In a Boltzmann Machine, only visible units are connected with data and hidden units are used to assist visible units to describe the distribution of data. Therefore, the model conceptually splits into the visible part and hidden part, while it still maintains a fully connected network among these units. 

``Stochastic'' is introduced for Boltzmann Machine to be improved from Hopfield Network regarding leaping out of the local optimum or oscillation of states. Inspired by physics, a method to transfer state regardless current energy is introduced: Set a state to State 1 (which means the state is on) regardless of the current state with the following probability: 
\begin{align*}
p = \dfrac{1}{1 + e^{-\frac{\Delta E}{T}}}
\end{align*}
where $\Delta E$ stands for the difference of energies when the state is on and off, i.e. $\Delta E = E_{s=1} - E_{s=0}$. $T$ stands for the temperature. The idea of $T$ is inspired by a physics process that the higher the temperature is, the more likely the state will transfer\footnote{Molecules move faster when more kinetic energy is provided, which could be achieved by heating.}. In addition, the probability of higher energy state transferring to lower energy state will be always greater than the reverse process\footnote{This corresponds to Zeroth Law of Thermodynamics.}. This idea is highly related to a very popular optimization algorithm called Simulated Annealing \citep{khachaturyan1979statistical,aarts1988simulated} back then, but Simulated Annealing is hardly relevant to nowadays deep learning society. Regardless of the historical importance that the term $T$ introduces, within this section, we will assume $T=1$ as a constant, for the sake of simplification. 

\subsubsection{Energy of Boltzmann Machine}
The energy function of Boltzmann Machine is defined the same as how Equation~\ref{eq:hopfield} is defined for Hopfield Network, except that now visible units and hidden units are noted separately, as following:
\begin{align*}
E(v,h) = -\sum_iv_ib_i - \sum_kh_kb_k - \sum_{i,j}v_iv_jw_{ij} - \sum_{i,k}v_ih_kw_{ik} - \sum_{k,l}h_kh_lw_{k,l}
\end{align*}
where $v$ stands for visible units, $h$ stands for hidden units. This equation also connects back to Equation~\ref{eq:spin}, except that Boltzmann Machine splits the energy function according to hidden units and visible units. 

Based on this energy function, the probability of a joint configuration over both visible unit the hidden unit can be defined as following:
\begin{align*}
p(v, h) = \dfrac{e^{-E(v,h)}}{\sum_{m,n}e^{-E(m,n)}}
\end{align*}
The probability of visible/hidden units can be achieved by marginalizing this joint probability.

For example, by marginalizing out hidden units, we can get the probability distribution of visible units:
\begin{align*}
p(v) = \dfrac{\sum_he^{-E(v,h)}}{\sum_{m,n}e^{-E(m,n)}}
\end{align*}
which could be used to sample visible units, i.e. generating data. 

When Boltzmann Machine is trained to its stable state, which is called \textbf{thermal equilibrium}, the distribution of these probabilities $p(v,h)$ will remain constant because the distribution of energy will be a constant. However, the probability for each visible unit or hidden unit may vary and the energy may not be at their minimum. This is related to how thermal equilibrium is defined, where the only constant factor is the distribution of each part of the system. 

\textbf{Thermal equilibrium} can be a hard concept to understand. One can imagine that pouring a cup of hot water into a bottle and then pouring a cup of cold water onto the hot water. At start, the bottle feels hot at bottom and feels cold at top and gradually the bottle feels mild as the cold water and hot water mix and heat is transferred. However, the temperature of the bottle becomes mild stably (corresponding to the distribution of $p(v,h)$) does not necessarily mean that the molecules cease to move (corresponding to each $p(v,h)$).  

\subsubsection{Parameter Learning}
The common way to train the Boltzmann machine is to determine the parameters that maximize the likelihood of the observed data. Gradient descent on the log of the likelihood function is usually performed to determine the parameters. For simplicity, the following derivation is based on a single observation. 

First, we have the log likelihood function of visible units as
\begin{align*}
l(v;w) = \log p(v;w) = \log \sum_h e^{-E_{v,h}} - \log \sum_{m,n}e^{-E_{m,n}}
\end{align*}
where the second term on RHS is the normalizer. 

Now we take the derivative of log likelihood function w.r.t $w$, and simplify it, we have:
\begin{align*}
\dfrac{\partial l(v;w)}{\partial w} =& -\sum_h p(h|v)\dfrac{\partial E(v,h)}{\partial w} + \sum_{m,n}p(m,n)\dfrac{\partial E(m,n)}{\partial w} \\
=&-\mathbb{E}_{p(h|v)}\dfrac{\partial E(v,h)}{\partial w} + \mathbb{E}_{p(m,n)}\dfrac{\partial E(m,n)}{\partial w}
\end{align*}
where $\mathbb{E}$ denotes expectation. 
Thus the gradient of the likelihood function is composed of two parts. The first part is expected gradient of the energy function with respect to the conditional distribution $p(h|v)$. The second part is expected gradient of the energy function with respect to the joint distribution over all variable states. However, calculating these expectations is generally infeasible for any realistically-sized model, as it involves summing over a huge number of possible states/configurations. The general approach for solving this problem is to use Markov Chain Monte Carlo (MCMC) to approximate these sums:
\begin{align}
\dfrac{\partial l(v;w)}{\partial w} = -<s_i,s_j>_{p(h_{data}|v_{data})} + <s_i,s_j>_{p(h_{model}|v_{model})}
\label{eq:bm,energy}
\end{align}
where $<\cdot>$ denotes expectation. 

Equation~\ref{eq:bm,energy} is the difference between the expectation value of product of states while the data is fed into visible states and the expectation of product of states while no data is fed. The first term is calculated by taking the average value of the energy function gradient when the visible and hidden units are being driven by observed data samples. In practice, this first term is generally straightforward to calculate. Calculating the second term is generally more complicated and involves running a set of Markov chains until they reach the current model’s equilibrium distribution, then taking the average energy function gradient based on those samples.

However, this sampling procedure could be very computationally complicated, which motivates the topic in next section, the Restricted Boltzmann Machine. 

\subsection{Restricted Boltzmann Machine}
Restricted Boltzmann Machine (RBM), originally known as Harmonium when invented by \citet{smolensky1986information}, is a version of Boltzmann Machine with a restriction that there is no connections either between visible units or between hidden units. 

\begin{figure}
\centering
\includegraphics[width=0.8\textwidth]{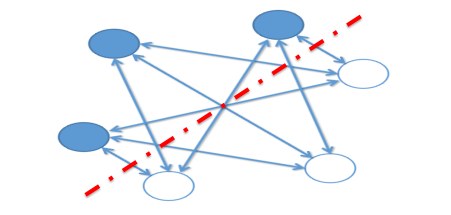}
\caption{Illustration of Restricted Boltzmann Machine. With the restriction that there is no connections between hidden units (shaded nodes) and no connections between visible units (unshaded nodes), the Boltzmann Machine turns into a Restricted Boltzmann Machine. The model now is a a bipartite graph. }
\label{fig:rbm}
\end{figure}

Figure~\ref{fig:rbm} is an illustration of how Restricted Boltzmann Machine is achieved based on Boltzmann Machine (Figure~\ref{fig:bm}): the connections between hidden units, as well as the connections between visible units are removed and the model becomes a bipartite graph. With this restriction introduced, the energy function of RBM is much simpler:
\begin{align}
E(v,h) = -\sum_iv_ib_i - \sum_kh_kb_k - \sum_{i,k}v_ih_kw_{ik}
\label{eq:rbm}
\end{align}

\subsubsection{Contrastive Divergence}
RBM can still be trained in the same way as how Boltzmann Machine is trained. Since the energy function of RBM is much simpler, the sampling method used to infer the second term in Equation~\ref{eq:bm,energy} becomes easier. Despite this relative simplicity, this learning procedure still requires a large amount of sampling steps to approximate the model distribution. 

To emphasize the difficulties of such a sampling mechanism, as well as to simplify follow-up introduction, we re-write Equation~\ref{eq:bm,energy} with a different set of notations, as following:
\begin{align}
\dfrac{\partial l(v;w)}{\partial w} = -<s_i,s_j>_{p_0} + <s_i,s_j>_{p_\infty}
\label{eq:cd}
\end{align}
here we use $p_0$ to denote data distribution and $p_\infty$ to denote model distribution. Other notations remain unchanged. Therefore, the difficulty of mentioned methods to learn the parameters is that it requires potentially ``infinitely'' many sampling steps to approximate the model distribution. 

\citet{hinton2002training} overcame this issue magically, with the introduction of a method named Contrastive Divergence. Empirically, he found that one does not have to perform ``infinitely'' many sampling steps to converge to the model distribution, a finite $k$ steps of sampling is enough. Therefore, Equation~\ref{eq:cd} is effectively re-written into:
\begin{align*}
\dfrac{\partial l(v;w)}{\partial w} = -<s_i,s_j>_{p_0} + <s_i,s_j>_{p_k}
\end{align*}
Remarkably, \citet{hinton2002training} showed that $k=1$ is sufficient for the learning algorithm to work well in practice. 

\citet{carreira2005contrastive} attempted to justify Contrastive Divergence in theory, but their derivation led to a negative conclusion that Contrastive Divergence is a biased algorithm, and a finite $k$ cannot represent the model distribution. However, their empirical results suggested that finite $k$ can approximate the model distribution well enough, resulting a small enough bias. In addition, the algorithm works well in practice, which strengthened the idea of Contrastive Divergence. 

With the reasonable modeling power and a fast approximation algorithm, RBM quickly draws great attention and becomes one of the most fundamental building blocks of deep neural networks. In the following two sections, we will introduce two distinguished deep neural networks that are built based on RBM/Boltzmann Machine, namely Deep Belief Nets and Deep Boltzmann Machine.  

\subsection{Deep Belief Nets}
Deep Belief Networks is introduced by \citet{hinton2006fast}\footnote{This paper is generally seen as the opening of nowadays Deep Learning era, as it first introduces the possibility of training a deep neural network by layerwise training}, when he showed that RBMs can be stacked and trained in a greedy manner. 

\begin{figure}
\centering
\includegraphics[width=0.8\textwidth]{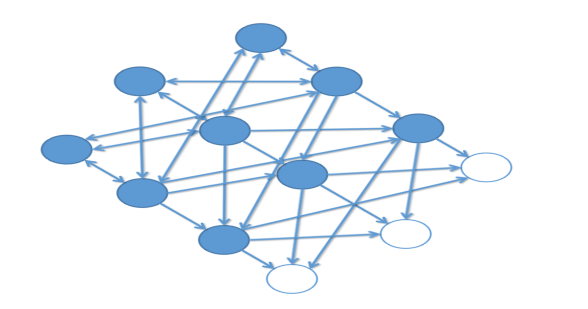}
\caption{Illustration of Deep Belief Networks. Deep Belief Networks is not just stacking RBM together. The bottom layers (layers except the top one) do not have the bi-directional connections, but only connections top down. }
\label{fig:dbn}
\end{figure}

Figure~\ref{fig:dbn} shows the structure of a three-layer Deep Belief Networks. Different from stacking RBM, DBN only allows bi-directional connections (RBM-type connections) on the top one layer while the following bottom layers only have top-down connections. Probably a better way to understand DBN is to think it as multi-layer generative models. Despite the fact that DBN is generally described as a stacked RBM, it is quite different from putting one RBM on the top of the other. It is probably more appropriate to think DBN as a one-layer RBM with extended layers specially devoted to generating patterns of data. 

Therefore, the model only needs to sample for the thermal equilibrium at the topmost layer and then pass the visible states top down to generate the data. 

\subsubsection{Parameter Learning}
Parameter learning of a Deep Belief Network falls into two steps: the first step is layer-wise pre-training and the second step is fine-tuning. 
\paragraph{Layerwise Pre-training} 
The success of Deep Belief Network is largely due to the introduction of the layer-wised pretraining. The idea is simple, but the reason why it works still attracts researchers. The pre-training is simply to first train the network component by component bottom up: treating the first two layers as an RBM and train it, then treat the second layer and third layer as another RBM and train for the parameters. 

Such an idea turns out to offer a critical support of the success of the later fine-tuning process. Several explanations have been attempted to explain the mechanism of pre-training:
\begin{itemize}
\item Intuitively, pre-training is a clever way of initialization. It puts the parameter values in the appropriate range for further fine-tuning. 
\item \cite{bengio2007greedy} suggested that unsupervised pre-training initializes the model to a point in parameter space which leads to a more effective optimization process, that the optimization can find a lower minimum of the empirical cost function. 
\item \cite{erhan2010does} empirically argued for a regularization explanation, that unsupervised pretraining guides the learning towards basins of attraction of minima that support better generalization from the training data set. 
\end{itemize}

In addition to Deep Belief Networks, this pretraining mechanism also inspires the pre-training for many other classical models, including the autoencoders \citep{poultney2006efficient,bengio2007greedy}, Deep Boltzmann Machines \citep{salakhutdinov2009deep} and some models inspired by these classical models like \citep{yu2010roles}. 

After the pre-training is performed, fine-tuning is carried out to further optimize the network to search for the parameters that lead to a lower minimum. For Deep Belief Networks, there are two different fine tuning strategies dependent on the goals of the network. 

\paragraph{Fine Tuning for Generative Model}
Fine-tuning for a generative model is achieved with a contrastive version of wake-sleep algorithm \citep{hinton1995wake}. This algorithm is intriguing for the reason that it is designed to interpret how the brain works. Scientists have found that sleeping is a critical process of brain function and it seems to be an inverse version of how we learn when we are awake. The wake-sleep algorithm also has two steps. In wake phase, we propagate information bottom up to adjust top-down weights for reconstructing the layer below. Sleep phase is the inverse of wake phase. We propagate the information top down to adjust bottom-up weights for reconstructing the layer above. 

The contrastive version of this wake-sleep algorithm is that we add one Contrastive Divergence phase between wake phase and sleep phase. The wake phase only goes up to the visible layer of the top RBM, then we sample the top RBM with Contrastive Divergence, then a sleep phase starts from the visible layer of top RBM. 

\paragraph{Fine Tuning for Discriminative Model}
The strategy for fine tuning a DBN as a discriminative model is to simply apply standard backpropagation to pre-trained model since we have labels of data. 
However, pre-training is still necessary in spite of the generally good performance of backpropagation. 

\subsection{Deep Boltzmann Machine}	
The last milestone we introduce in the family of deep generative model is Deep Boltzmann Machine introduced by \cite{salakhutdinov2009deep}. 

\begin{figure}
\centering
\includegraphics[width=0.8\textwidth]{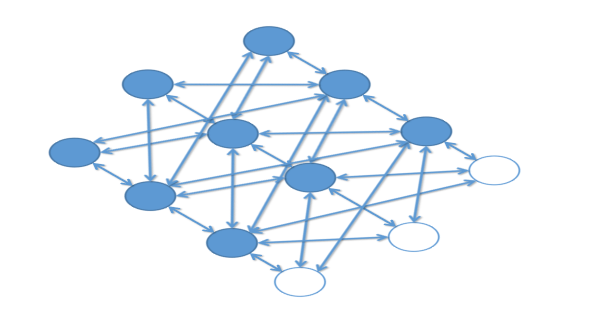}
\caption{Illustration of Deep Boltzmann Machine. Deep Boltzmann Machine is more like stacking RBM together. Connections between every two layers are bidirectional.}
\label{fig:dbm}
\end{figure}

Figure~\ref{fig:dbm} shows a three layer Deep Boltzmann Machine (DBM). The distinction between DBM and DBN mentioned in the previous section is that DBM allows bidirectional connections in the bottom layers. Therefore, DBM represents the idea of stacking RBMs in a much better way than DBN, although it might be clearer if DBM is named as Deep Restricted Boltzmann Machine. 

Due to the nature of DBM, its energy function is defined as an extension of the energy function of an RBM (Equation~\ref{eq:rbm}), as showed in the following:
\begin{align*}
E(v,h) = -\sum_iv_ib_i - \sum_{n=1}^{N}\sum_kh_{n,k}b_{n,k} - \sum_{i,k}v_iw_{ik}h_k - \sum_{n=1}^{N-1}\sum_{k,l}h_{n,k}w_{n,k,l}h_{n+1, l}
\end{align*}
for a DBM with $N$ hidden layers. 

This similarity of energy function grants the possibility of training DBM with constrative divergence. However, pre-training is typically necessary. 

\subsubsection{Deep Boltzmann Machine (DBM) v.s. Deep Belief Networks (DBN)}
As their acronyms suggest, Deep Boltzmann Machine and Deep Belief Networks have many similarities, especially from the first glance. Both of them are deep neural networks originates from the idea of Restricted Boltzmann Machine. (The name ``Deep Belief Network'' seems to indicate that it also partially originates from Bayesian Network \citep{krieg2001tutorial}.)
Both of them also rely on layerwise pre-training for a success of parameter learning. 

However, the fundamental differences between these two models are dramatic, introduced by how the connections are made between bottom layers (un-directed/bi-directed v.s. directed). The bidirectional structure of DBM grants the possibility of DBM to learn a more complex pattern of data. It also grants the possibility for the approximate inference procedure to incorporate top-down feedback in addition to an initial bottom-up pass, allowing Deep Boltzmann Machines to better propagate uncertainty about ambiguous inputs. 

\subsection{Deep Generative Models: Now and the Future}
Deep Boltzmann Machine is the last milestone we discuss in the history of generative models, but there are still much work after DBM and even more to be done in the future. 

\cite{lake2015human} introduces a Bayesian Program Learning framework that can simulate human learning abilities with large scale visual concepts. In addition to its performance on one-shot learning classification task, their model passes the visual Turing Test in terms of generating handwritten characters from the world’s alphabets. In other words, the generative performance of their model is indistinguishable from human's behavior. Being not a deep neural model itself, their model outperforms several concurrent deep neural networks. Deep neural counterpart of the Bayesian Program Learning framework can be surely expected with even better performance.  

Conditional image generation (given part of the image) is also another interesting topic recently. The problem is usually solved by Pixel Networks (Pixel CNN \citep{van2016conditional} and Pixel RNN \citep{oord2016pixel}). However, given a part of the image seems to simplify the generation task. 

Another contribution to generative models is Generative Adversarial Network \citep{goodfellow2014generative}, however, GAN is still too young to be discussed in this paper. 

\newpage
\section{Convolutional Neural Networks and Vision Problems}
\label{sec:cnn}
In this section, we will start to discuss a different family of models: the Convolutional Neural Network (CNN) family. Distinct from the family in the previous section, Convolutional Neural Network family mainly evolves from the knowledge of human visual cortex. Therefore, in this section, we will first introduce one of the most important reasons that account for the success of convolutional neural networks in vision problems: its bionic design to replicate human vision system. The nowadays convolutional neural networks probably originate more from the such a design rather than from the early-stage ancestors. With these background set-up, we will then briefly introduce the successful models that make themselves famous through the ImageNet Challenge \citep{deng2009imagenet}. At last, we will present some known problems of the vision task that may guide the future research directions in vision tasks.

\subsection{Visual Cortex}
Convolutional Neural Network is widely known as being inspired by visual cortex, however, except that some publications discuss this inspiration briefly \citep{poggio2013models,cox2014neural}, few resources present this inspiration thoroughly. In this section, we focus on the discussion about basics on visual cortex \citep{hubel1959receptive}, which lays the ground for further study in Convolutional Neural Networks. 

The visual cortex of the brain, located in the occipital lobe which is located at the back of the skull, is a part of the cerebral cortex that plays an important role in processing visual information. Visual information coming from the eye, goes through a series of brain structures and reaches the visual cortex. The parts of the visual cortex that receive the sensory inputs is known as the primary visual cortex, also known as area V1. Visual information is further managed by extrastriate areas, including visual areas two (V2) and four (V4). There are also other visual areas (V3, V5, and V6), but in this paper, we primarily focus on the visual areas that are related to object recognition, which is known as ventral stream and consists of areas V1, V2, V4 and inferior temporal gyrus, which is one of the higher levels of the ventral stream of visual processing, associated with the representation of complex object features, such as global shape, like face perception \citep{haxby2000distributed}. 

\begin{figure}
\centering
\includegraphics[width=0.7\textwidth]{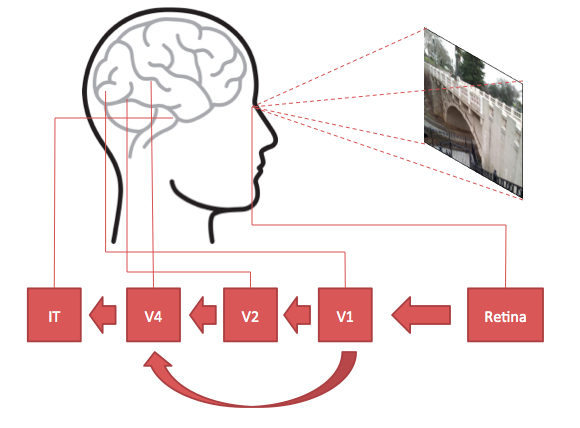}
\caption{A brief illustration of ventral stream of the visual cortex in human vision system. It consists of primary visual cortex (V1), visual areas (V2 and V4) and  inferior temporal gyrus. }
\label{fig:vc}
\end{figure}

Figure~\ref{fig:vc} is an illustration of the ventral stream of the visual cortex. It shows the information process procedure from the retina which receives the image information and passes all the way to inferior temporal gyrus. For each component:
\begin{itemize}
\item Retina converts the light energy that comes from the rays bouncing off of an object into chemical energy. This chemical energy is then converted into action potentials that are transferred onto primary visual cortex. (In fact, there are several other brain structures involved between retina and V1, but we omit these structures for simplicity\footnote{We deliberately discuss the components that have connections with established technologies in convolutional neural network, one who is interested in developing more powerful models is encouraged to investigate other components. }.)
\item Primary visual cortex (V1) mainly fulfills the task of edge detection, where an edge is an area with strongest local contrast in the visual signals. 
\item V2, also known as secondary visual cortex, is the first region within the visual association area. It receives strong feedforward connections from V1 and sends strong connections to later areas. In V2, cells are tuned to extract mainly simple properties of the visual signals such as orientation, spatial frequency, and colour, and a few more complex properties. 
\item V4 fulfills the functions including detecting object features of intermediate complexity, like simple geometric shapes, in addition to orientation, spatial frequency, and color. V4 is also shown with strong attentional modulation \citep{moran1985selective}. V4 also receives direct input from V1. 
\item Inferior temporal gyrus (TI) is responsible for identifying the object based on the color and form of the object and comparing that processed information to stored memories of objects to identify that object \citep{kolb2014introduction}. In other words, IT performs the semantic level tasks, like face recognition. 
\end{itemize}

Many of the descriptions of functions about visual cortex should revive a recollection of convolutional neural networks for the readers that have been exposed to some relevant technical literature. Later in this section, we will discuss more details about convolutional neural networks, which will help build explicit connections. Even for readers that barely have knowledge in convolutional neural networks, this hierarchical structure of visual cortex should immediately ring a bell about neural networks.   

Besides convolutional neural networks, visual cortex has been inspiring the works in computer vision for a long time. For example, \cite{li1998neural} built a neural model inspired by the primary visual cortex (V1). In another granularity, \cite{serre2005object} introduced a system with feature detections inspired from the visual cortex. \cite{de2012models} published a book describing the models of information processing in the visual cortex. \cite{poggio2013models} conducted a more comprehensive survey on the relevant topic, but they didn't focus on any particular subject in detail in their survey. In this section, we discuss the connections between visual cortex and convolutional neural networks in details. We will begin with Neocogitron, which borrows some ideas from visual cortex and later inspires convolutional neural network. 

\subsection{Neocogitron and Visual Cortex}
Neocogitron, proposed by \cite{fukushima1980neocognitron}, is generally seen as the model that inspires Convolutional Neural Networks on the computation side. It is a neural network that consists of two different kinds of layers (S-layer as feature extractor and C-layer as structured connections to organize the extracted features.) 

S-layer consists of a number of S-cells that are inspired by the cell in primary visual cortex. It serves as a feature extractor. Each S-cell can be ideally trained to be responsive to a particular feature presented in its receptive field. Generally, local features such as edges in particular orientations are extracted in lower layers while global features are extracted in higher layers. This structure highly resembles how human conceive objects. C-layer resembles complex cell in the higher pathway of visual cortex. It is mainly introduced for shift invariant property of features extracted by S-layer. 

\subsubsection{Parameter Learning}
During parameter learning process, only the parameters of S-layer are updated. Neocogitron can also be trained unsupervisedly, for a good feature extractor out of S-layers. The training process for S-layer is very similar to Hebbian Learning rule, which strengthens the connections between S-layer and C-layer for whichever S-cell shows the strongest response. This training mechanism also introduces the problem Hebbian Learning rule introduces, that the strength of connections will saturate (since it keeps increasing). The solution was also introduced by \cite{fukushima1980neocognitron}, which was introduced with the name ``inhibitory cell''. It performed the function as a normalization to avoid the problem. 

\subsection{Convolutional Neural Network and Visual Cortex}
Now we proceed from Neocogitron to Convolutional Neural Network. First, we will introduce the building components: convolutional layer and subsampling layer. Then we assemble these components to present Convolutional Neural Network, using LeNet as an example. 
\subsubsection{Convolution Operation}
Convolution operation is strictly just a mathematical operation, which should be treated equally with other operations like addition or multiplication and should not be discussed particularly in a machine learning literature. However, we still discuss it here for completeness and for the readers who may not be familiar with it. 

Convolution is a mathematical operation on two functions (e.g. $f$ and $g$) and produces a third function $h$, which is an integral that expresses the amount of overlap of one function ($f$) as it is shifted over the other function ($g$). It is described formally as the following:
\begin{align*}
h(t) = \int_{-\infty}^\infty f(\tau)g(t-\tau)d\tau
\end{align*}
and denoted as $h=f\star g$. 

\begin{figure}
\centering
\includegraphics[width=0.7\textwidth]{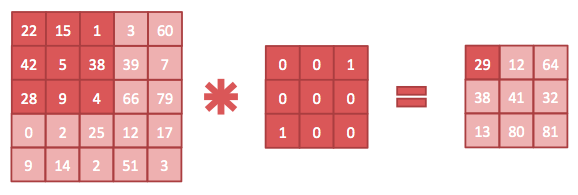}
\caption{A simple illustration of two dimension convolution operation. }
\label{fig:conv}
\end{figure}

Convolutional neural network typically works with two-dimensional convolution operation, which could be summarized in Figure~\ref{fig:conv}.

As showed in Figure~\ref{fig:conv}, the leftmost matrix is the input matrix. The middle one is usually called a kernel matrix. Convolution is applied to these matrices and the result is showed as the rightmost matrix. The convolution process is an element-wise product followed by a sum, as showed in the example. When the left upper $3\times3$ matrix is convoluted with the kernel, the result is $29$. Then we slide the target $3\times3$ matrix one column right, convoluted with the kernel and get the result $12$. We keep sliding and record the results as a matrix. Because the kernel is $3\times3$, every target matrix is $3\times3$, thus, every $3\times3$ matrix is convoluted to one digit and the whole $5\times5$ matrix is shrunk into $3\times3$ matrix. (Because $5 - (3-1) = 3$. The first $3$ means the size of the kernel matrix. ) 

One should realize that convolution is locally shift invariant, which means that for many different combinations of how the nine numbers in the upper $3\times 3$ matrix are placed, the convoluted result will be $29$. This invariant property plays a critical role in vision problem because that in an ideal case, the recognition result should not be changed due to shift or rotation of features. This critical property is used to be solved elegantly by \cite{lowe1999object,bay2006surf}, but convolutional neural network brought the performance up to a new level. 

\subsubsection{Connection between CNN and Visual Cortex}
With the ideas about two dimension convolution, we further discuss how convolution is a useful operation that can simulate the tasks performed by visual cortex. 

The convolution operation is usually known as kernels. By different choices of kernels, different operations of the images could be achieved. Operations are typically including identity, edge detection, blur, sharpening etc. By introducing random matrices as convolution operator, some interesting properties might be discovered. 

\begin{figure}
\subfloat[Identity kernel]{\includegraphics[width=0.45\textwidth]{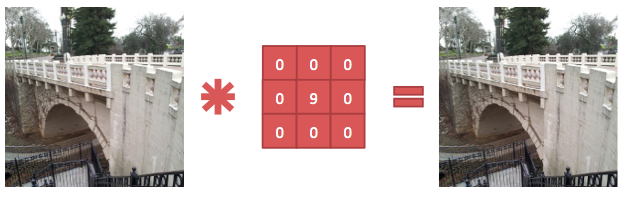}}
\hfill
\subfloat[Edge detection kernel]{\includegraphics[width=0.45\textwidth]{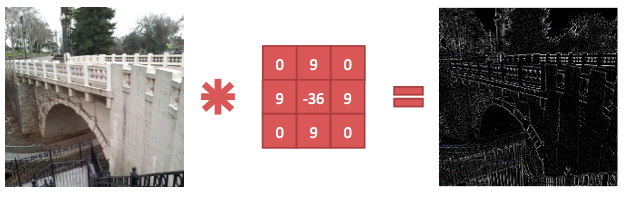}}
\hfill
\subfloat[Blur kernel]{\includegraphics[width=0.45\textwidth]{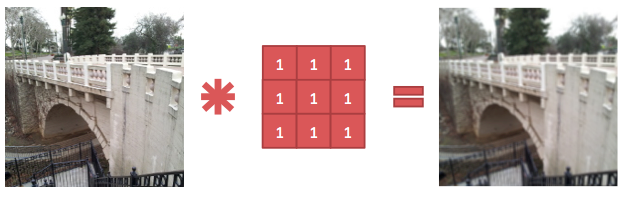}}
\hfill
\subfloat[Sharpen kernel]{\includegraphics[width=0.45\textwidth]{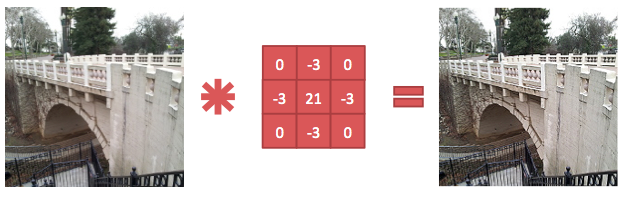}}
\hfill
\subfloat[Lighten kernel]{\includegraphics[width=0.45\textwidth]{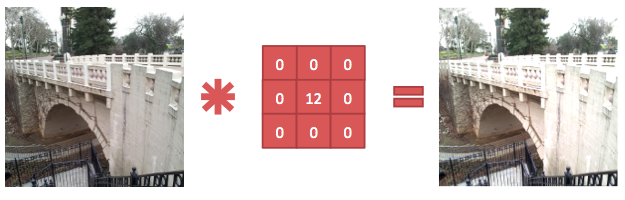}}
\hfill
\subfloat[Darken kernel]{\includegraphics[width=0.45\textwidth]{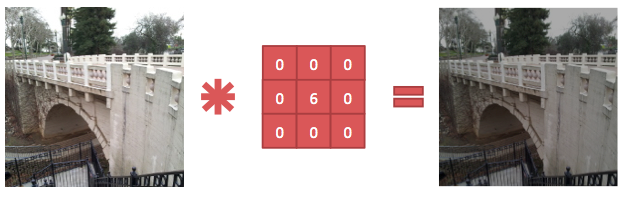}}
\hfill
\subfloat[Random kernel 1]{\includegraphics[width=0.45\textwidth]{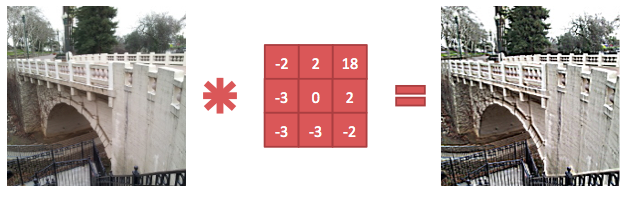}}
\hfill
\subfloat[Random kernel 2]{\includegraphics[width=0.45\textwidth]{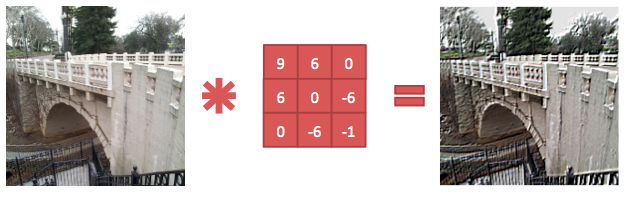}}
\caption{Convolutional kernels example. Different kernels applied to the same image will result in differently processed images. Note that there is a $\frac{1}{9}$ divisor applied to these kernels.}
\label{fig:kernel}
\end{figure}

Figure~\ref{fig:kernel} is an illustration of some example kernels that are applied to the same figure. One can see that different kernels can be applied to fulfill different tasks. Random kernels can also be applied to transform the image into some interesting outcomes. 

Figure~\ref{fig:kernel} (b) shows that edge detection, which is one of the central tasks of primary visual cortex, can be fulfilled by a clever choice of kernels. Furthermore, clever selection of kernels can lead us to a success replication of visual cortex. As a result, learning a meaningful convolutional kernel (i.e. parameter learning) is one of the central tasks in convolutional neural networks when applied to vision tasks. This also explains that why many well-trained popular models can usually perform well in other tasks with only limited fine-tuning process: the kernels have been well trained and can be universally applicable. 

With the understanding of the essential role convolution operation plays in vision tasks, we proceed to investigate some major milestones along the way. 

\subsection{The Pioneer of Convolutional Neural Networks: LeNet}
This section is devoted to a model that is widely recognized as the first convolutional neural network: LeNet, invented by \cite{le1990handwritten} (further made popular with \citep{lecun1998gradient}). It is inspired from the Neocogitron. In this section, we will introduce convolutional neural network via introducing LeNet. 

\begin{figure}
\centering
\includegraphics[width=0.8\textwidth]{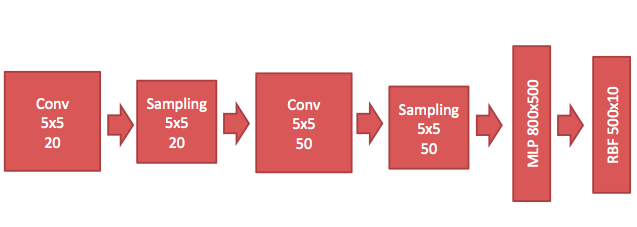}
\caption{An illustration of LeNet, where Conv stands for convolutional layer and Sampling stands for SubSampling Layer.}
\label{fig:lenet}
\end{figure}

Figure~\ref{fig:lenet} shows an illustration of the architecture of LeNet. It consists of two pairs of Convolutional Layer and Subsampling Layer and is further connected with fully connected layer and an RBF layer for classification. 

\subsubsection{Convolutional Layer}
A convolutional layer is primarily a layer that performs convolution operation. As we have discussed previously, a clever selection of convolution kernel can effectively simulate the task of visual cortex. Convolutional layer introduces another operation after convolution to assist the simulation to be more successful: the non-linearity transform. 

Considering a ReLU \citep{nair2010rectified} non-linearity transform, which is defined as following:
\begin{align*}
f(x) = \max(0, x)
\end{align*}
which is a transform that removes the negative part of the input, resulting in a clearer contrast of meaningful features as opposed to other side product the kernel produces. 

Therefore, this non-linearity grants the convolution more power in extracting useful features and allows it to simulate the functions of visual cortex more closely. 

\subsubsection{Subsampling Layer}
Subsampling Layer performs a simpler task. It only samples one input out every region it looks into. Some different strategies of sampling can be considered, like max-pooling (taking the maximum value of the input), average-pooling (taking the averaged value of input) or even probabilistic pooling (taking a random one.) \citep{lee2009convolutional}. 

Sampling turns the input representations into smaller and more manageable embeddings. More importantly, sampling makes the network invariant to small transformations, distortions, and translations in the input image. A small distortion in the input will not change the outcome of pooling since we take the maximum/average value in a local neighborhood. 

\subsubsection{LeNet}
With the two most important components introduced, we can stack them together to assemble a convolutional neural network. Following the recipe of Figure~\ref{fig:lenet}, we will end up with the famous LeNet. 

LeNet is known as its ability to classify digits and can handle a variety of different problems of digits including variances in position and scale, rotation and squeezing of digits, and even different stroke width of the digit. Meanwhile, with the introduction of LeNet, \cite{lecun1998mnist} also introduces the MNIST database, which later becomes the standard benchmark in digit recognition field. 

\subsection{Milestones in ImageNet Challenge}
With the success of LeNet, Convolutional Neural Network has been shown with great potential in solving vision tasks. These potentials have attracted a large number of researchers aiming to solve vision task regarding object recognition in CIFAR classification \citep{krizhevsky2009learning} and ImageNet challenge \citep{russakovsky2015imagenet}. Along with this path, several superstar milestones have attracted great attentions and has been applied to other fields with good performance. In this section, we will briefly discuss these models. 

\subsubsection{AlexNet}
While LeNet is the one that starts the era of convolutional neural networks, AlexNet, invented by \cite{krizhevsky2012imagenet}, is the one that starts the era of CNN used for ImageNet classification. AlexNet is the first evidence that CNN can perform well on this historically difficult ImageNet dataset and it performs so well that leads the society into a competition of developing CNNs. 

\begin{figure}
\centering
\includegraphics[width=0.8\textwidth]{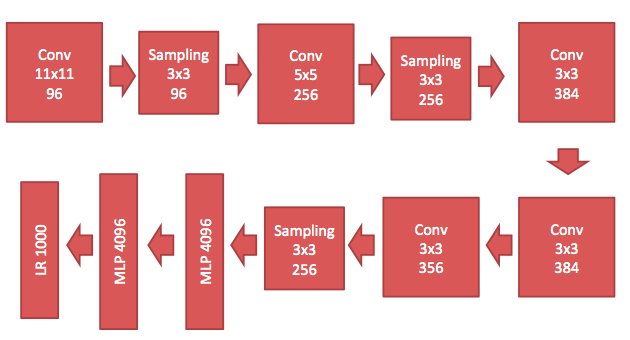}
\caption{An illustration of AlexNet}
\label{fig:alexnet}
\end{figure}

The success of AlexNet is not only due to this unique design of architecture but also due to the clever mechanism of training. To avoid the computationally expensive training process, AlexNet has been split into two streams and trained on two GPUs. It also used data augmentation techniques that consist of image translations, horizontal reflections, and patch extractions.

The recipe of AlexNet is shown in Figure~\ref{fig:alexnet}. However, rarely any lessons can be learned from the architecture of AlexNet despite its remarkable performance. Even more unfortunately, the fact that this particular architecture of AlexNet does not have a well-grounded theoretical support pushes many researchers to blindly burn computing resources to test for a new architecture. Many models have been introduced during this period, but only a few may be worth mentioning in the future. 

\subsubsection{VGG}
In the blind competition of exploring different architectures, \cite{simonyan2014very} showed that simplicity is a promising direction with a model named VGG. Although VGG is deeper (19 layer) than other models around that time, the architecture is extremely simplified: all the layers are $3\times 3$ convolutional layer with a $2\times2$ pooling layer. This simple usage of convolutional layer simulates a larger filter while keeping the benefits of smaller filter sizes, because the combination of two $3\times3$ convolutional layers has an effective receptive field of a $5\times5$ convolutional layer, but with fewer parameters. 

The spatial size of the input volumes at each layer will decrease as a result of the convolutional and pooling layers, but the depth of the volumes increases because of the increased number of filters (in VGG, the number of filters doubles after each pooling layer). This behavior reinforces the idea of VGG to shrink spatial dimensions, but grow depth. 

VGG is not the winner of the ImageNet competition of that year (The winner is GoogLeNet invented by \cite{szegedy2015going}). GoogLeNet introduced several important concepts like Inception module and the concept later used by R-CNN \citep{girshick2014rich,girshick2015fast,ren2015faster}, but the arbitrary/creative design of architecture barely contribute more than what VGG does to the society, especially considering that Residual Net, following the path of VGG, won the ImageNet challenge in an unprecedented level. 

\subsubsection{Residual Net}
Residual Net (ResNet) is a 152 layer network, which was ten times deeper than what was usually seen during the time when it was invented by \cite{he2015deep}. Following the path VGG introduces, ResNet explores deeper structure with simple layer. However, naively increasing the number of layers will only result in worse results, for both training cases and testing cases \citep{he2015deep}. 

\begin{figure}
\centering
\includegraphics[width=0.3\textwidth]{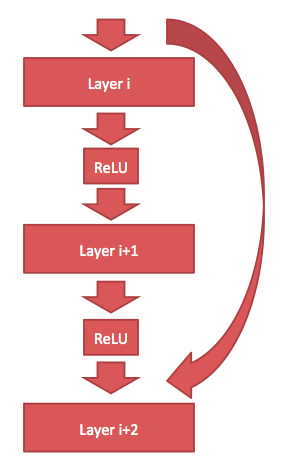}
\caption{An illustration of Residual Block of ResNet}
\label{fig:resnet}
\end{figure}

The breakthrough ResNet introduces, which allows ResNet to be substantially deeper than previous networks, is called Residual Block. The idea behind a Residual Block is that some input of a certain layer (denoted as $x$) can be passed to the component two layers later either following the traditional path which involves convolutional layers and ReLU transform succession (we denote the result as $f(x)$), or going through an express way that directly passes $x$ there. As a result, the input to the component two layers later is $f(x)+x$ instead of what is typically seen as $f(x)$. The idea of Residual Block is illustrated in Figure~\ref{fig:resnet}. 

In a complementary work, \cite{he2016identity} validated that residual blocks are essential for propagating information smoothly, therefore simplifies the optimization. They also extended the ResNet to a 1000-layer version with success on CIFAR data set. 

Another interesting perspective of ResNet is provided by \citep{veit2016residual}. They showed that ResNet behave behaves like ensemble of shallow networks: the express way introduced allows ResNet to perform as a collection of independent networks, each network is significantly shallower than the integrated ResNet itself. This also explains why gradient can be passed through the ultra-deep architecture without being vanished. (We will talk more about vanishing gradient problem when we discuss recurrent neural network in the next section.) Another work, which is not directly relevant to ResNet, but may help to understand it, is conducted by \cite{hariharan2015hypercolumns}. They showed that features from lower layers are informative in addition to what can be summarized from the final layer. 

ResNet is still not completely vacant from clever designs. The number of layers in the whole network and the number of layers that Residual Block allows identity to bypass are still choices that require experimental validations. Nonetheless, to some extent, ResNet has shown that critical reasoning can help the development of CNN better than blind experimental trails. In addition, the idea of Residual Block has been found in the actual visual cortex (In the ventral stream of the visual cortex, V4 can directly accept signals from primary visual cortex), although ResNet is not designed according to this in the first place.  

With the introduction of these state-of-the-art neural models that are successful in these challenges, \cite{canziani2016analysis} conducted a comprehensive experimental study comparing these models. Upon comparison, they showed that there is still room for improvement on fully connected layers that show strong inefficiencies for smaller batches of images. 

\subsection{Challenges and Chances for Fundamental Vision Problems}
ResNet is not the end of the story. New models and techniques appear every day to push the limit of CNNs further. For example, \cite{zhang2016residual} took a step further and put Residual Block inside Residual Block. \cite{zagoruyko2016wide} attempted to decrease the depth of network by increasing the width. However, incremental works of this kind are not in the scope of this paper. 

We would like to end the story of Convolutional Neural Networks with some of the current challenges of fundamental vision problems that may not able to be solved naively by investigation of machine learning techniques. 

\subsubsection{Network Property and Vision Blindness Spot}
Convolutional Neural Networks have reached to an unprecedented accuracy in object detection. However, it may still be far from industry reliable application due to some intriguing properties found by \cite{szegedy2013intriguing}. 

\cite{szegedy2013intriguing} showed that they could force a deep learning model to misclassify an image simply by adding perturbations to that image. More importantly, these perturbations may not even be observed by naked human eyes. In other words, two objects that look almost the same to human, may be recognized as different objects by a well-trained neural network (for example, AlexNet). They have also shown that this property is more likely to be a modeling problem, in contrast to problems raised by insufficient training. 

On the other hand, \cite{nguyen2015deep} showed that they could generate patterns that convey almost no information to human, being recognized as some objects by neural networks with high confidence (sometimes more than 99\%). Since neural networks are typically forced to make a prediction, it is not surprising to see a network classify a meaningless patter into something, however, this high confidence may indicate that the fundamental differences between how neural networks and human learn to know this world. 

\begin{figure}
\subfloat[]{\includegraphics[width=0.15\textwidth]{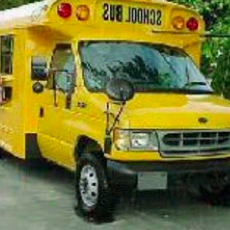}}
\hfill
\subfloat[]{\includegraphics[width=0.15\textwidth]{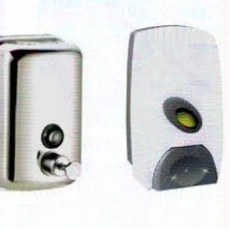}}
\hfill
\subfloat[]{\includegraphics[width=0.15\textwidth]{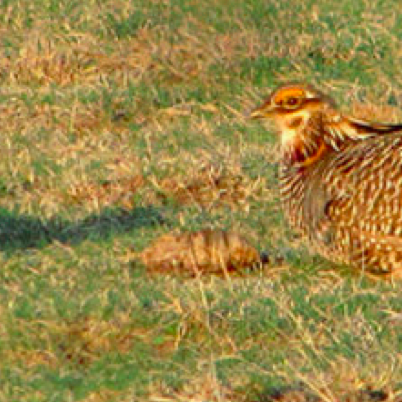}}
\hfill
\subfloat[]{\includegraphics[width=0.15\textwidth]{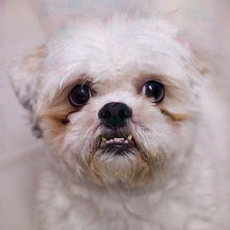}}
\\
\subfloat[]{\includegraphics[width=0.15\textwidth]{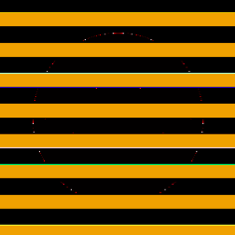}}
\hfill
\subfloat[]{\includegraphics[width=0.15\textwidth]{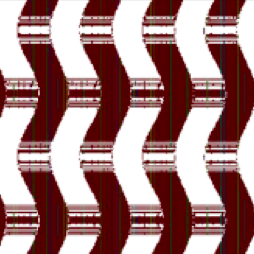}}
\hfill
\subfloat[]{\includegraphics[width=0.15\textwidth]{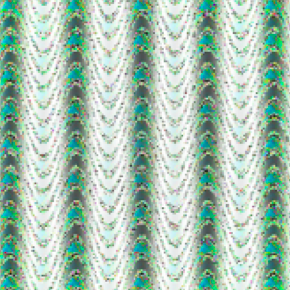}}
\hfill
\subfloat[]{\includegraphics[width=0.15\textwidth]{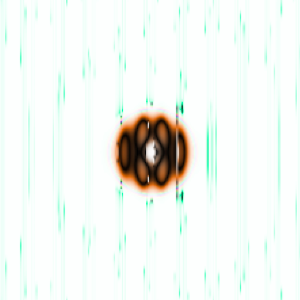}}
\caption{Illustrations of some mistakes of neural networks. (a)-(d) (from \citep{szegedy2013intriguing}) are adversarial images that are generated based on original images. The differences between these and the original ones are un-observable by naked eye, but the neural network can successfully classify original ones but fail adversarial ones. (e)-(h) (from \citep{nguyen2015deep}) are patterns that are generated. A neural network classify them into (e) school bus, (f) guitar, (g) peacock and (h) Pekinese respectively. }
\label{fig:fake}
\end{figure}

Figure~\ref{fig:fake} shows some examples from the aforementioned two works. With construction, we can show that the neural networks may misclassify an object, which should be easily recognized by the human, to something unusual. On the other hand, a neural network may also classify some weird patterns, which are not believed to be objects by the human, to something we are familiar with. Both of these properties may restrict the usage of deep learning to real world applications when a reliable prediction is necessary. 

Even without these examples, one may also realize that the reliable prediction of neural networks could be an issue due to the fundamental property of a matrix: the existence of null space. As long as the perturbation happens within the null space of a matrix, one may be able to alter an image dramatically while the neural network still makes the misclassification with high confidence. Null space works like a blind spot to a matrix and changes within null space are never sensible to the corresponding matrix. 

This blind spot should not discourage the promising future of neural networks. On the contrary, it makes the convolutional neural network resemble the human vision system in a deeper level. In the human vision system, blind spots \citep{gregory2011blind} also exist \citep{wandell1995foundations}. Interesting work might be seen about linking the flaws of human vision system to the defects of neural networks and helping to overcome these defects in the future. 

\subsubsection{Human Labeling Preference}
At the very last, we present some of the misclassified images of ResNet on ImageNet Challenge. Hopefully, some of these examples could inspire some new methodologies invented for the fundamental vision problem. 

\begin{figure}
\subfloat[flute]{\includegraphics[width=0.15\textwidth]{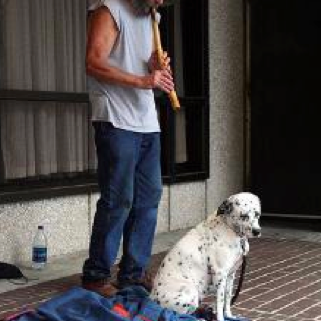}}
\hfill
\subfloat[guinea pig]{\includegraphics[width=0.15\textwidth]{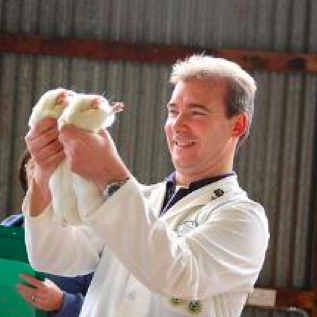}}
\hfill
\subfloat[wig]{\includegraphics[width=0.15\textwidth]{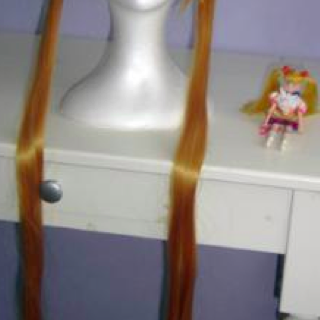}}
\hfill
\subfloat[seashore]{\includegraphics[width=0.15\textwidth]{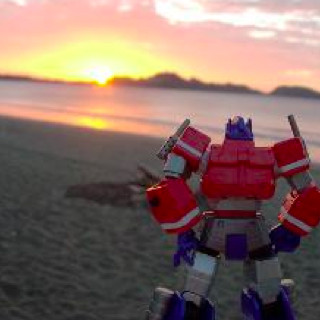}}
\\
\subfloat[alp]{\includegraphics[width=0.15\textwidth]{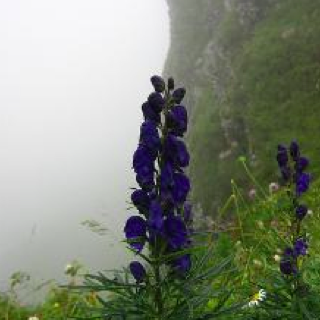}}
\hfill
\subfloat[screwdriver]{\includegraphics[width=0.15\textwidth]{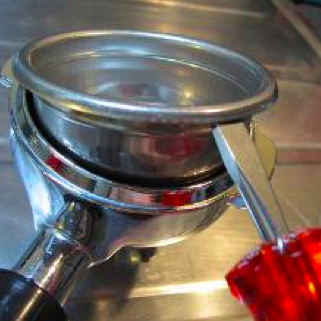}}
\hfill
\subfloat[comic book]{\includegraphics[width=0.15\textwidth]{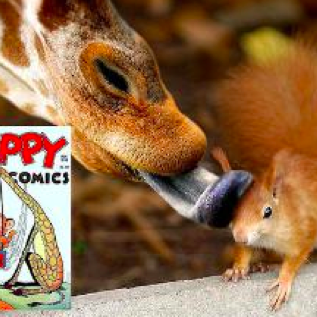}}
\hfill
\subfloat[sunglass]{\includegraphics[width=0.15\textwidth]{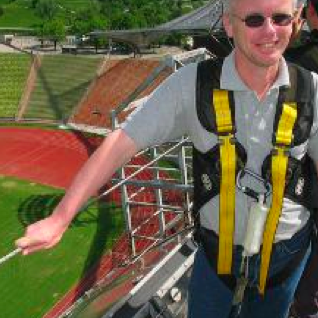}}
\caption{Some failed images of ImageNet classification by ResNet and the primary label associated with the image. }
\label{fig:imagenet}
\end{figure}

Figure~\ref{fig:imagenet} shows some misclassified images of ResNet when applied to ImageNet Challenge. These labels, provided by human effort, are very unexpected even to many other humans. Therefore, the 3.6\% error rate of ResNet (a general human usually predicts with error rate 5\%-10\%) is probably hitting the limit since the labeling preference of an annotator is harder to predict than the actual labels. For example, Figure~\ref{fig:imagenet} (a),(b),(h) are labeled as a tiny part of the image, while there are more important contents expressed by the image. On the other hand, Figure~\ref{fig:imagenet} (d) (e) are annotated as the background of the image while that image is obviously centering on other object. 

To further improve the performance ResNet reached, one direction might be to modeling the annotators' labeling preference. One assumption could be that annotators prefer to label an image to make it distinguishable. Some established work to modeling human factors \citep{wilson2015human} could be helpful. 

However, the more important question is that whether it is worth optimizing the model to increase the testing results on ImageNet dataset, since remaining misclassifications may not be a result of the incompetency of the model, but problems of annotations. 

The introduction of other data sets, like COCO \citep{lin2014microsoft}, Flickr \citep{plummer2015flickr30k}, and VisualGenome \citep{krishna2016visual} may open a new era of vision problems with more competitive challenges. However, the fundamental problems and experiences that this section introduces should never be forgotten.

\newpage
\section{Time Series Data and Recurrent Networks}
\label{sec:rnn}
In this section, we will start to discuss a new family of deep learning models that have attracted many attentions, especially for the tasks on time series data, or sequential data. The Recurrent Neural Network (RNN) is a class of neural network whose connections of units form a directed cycle; this nature grants its ability to work with temporal data. It has also been discussed in literature like \citep{grossberg2013recurrent} and \citep{lipton2015critical}. In this paper, we will continue to offer complementary views to other surveys with an emphasis on the evolutionary history of the milestone models and aim to provide insights into the future direction of coming models.  

\subsection{Recurrent Neural Network: Jordan Network and Elman Network}
As we have discussed previously, Hopfield Network is widely recognized as a recurrent neural network, although its formalization is distinctly different from how recurrent neural network is defined nowadays. Therefore, despite that other literature tend to begin the discussion of RNN with Hopfield Network, we will not treat it as a member of RNN family to avoid unnecessary confusion. 

The modern definition of ``recurrent'' is initially introduced by  \cite{jordan1986serial} as:
\begin{quotation}
If a network has one or more cycles, that is, if it is possible to follow a path from a unit back to itself, then the network is referred to as \textit{recurrent}. A \textit{nonrecurrent} network has no cycles. 
\end{quotation}
His model in \citep{jordan1986serial} is later referred to as Jordan Network. For a simple neural network with one hidden layer, with input denoted as $X$, weights of hidden layer denoted as $w_h$ and weights of output layer denoted as $w_y$, weights of recurrent computation denoted as $w_r$, hidden representation denoted as $h$ and output denoted as $y$, Jordan Network can be formulated as
\begin{align*}
h^{t} &= \sigma(W_hX+W_ry^{t-1}) \\
y &= \sigma(W_yh^{t})
\end{align*}

\begin{figure}
\centering
\subfloat[Structure of Jordan Network]{
  \includegraphics[clip,width=0.4\textwidth]{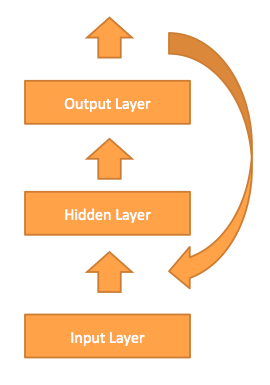}
}
\qquad
\subfloat[Structure of Elman Network]{
  \includegraphics[clip,width=0.4\textwidth]{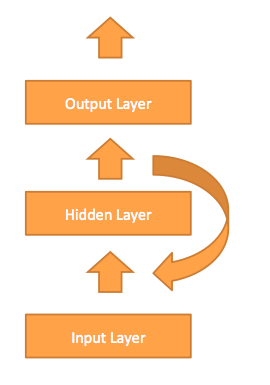}
}
\caption{The difference of recurrent structure from Jordan Network and Elman Network.}
\label{fig:rnn}
\end{figure}

A few years later, another RNN was introduced by \cite{elman1990finding}, when he formalized the recurrent structure slightly differently. Later, his network is known as Elman Network. Elman network is formalized as following:
\begin{align*}
h^{t} &= \sigma(W_hX+W_rh^{t-1}) \\
y &= \sigma(W_yh^{t})
\end{align*}
The only difference is that whether the information of previous time step is provided by previous output or previous hidden layer. This difference is further illustrated in Figure~\ref{fig:rnn}. The difference is illustrated to respect the historical contribution of these works. One may notice that there is no fundamental difference between these two structures since $y_t=W_yh_t$, therefore, the only difference lies in the choice of $W_r$. (Originally, Elman only introduces his network with $W_r=\mathbf{I}$, but more general cases could be derived from there.)

Nevertheless, the step from Jordan Network to Elman Network is still remarkable as it introduces the possibility of passing information from hidden layers, which significantly improve the flexibility of structure design in later work. 

\subsubsection{Backpropagation through Time}
The recurrent structure makes traditional backpropagation infeasible because of that with the recurrent structure, there is not an end point where the backpropagation can stop. 

Intuitively, one solution is to unfold the recurrent structure and expand it as a feedforward neural network with certain time steps and then apply traditional backpropagation onto this unfolded neural network. This solution is known as Backpropagation through Time (BPTT), independently invented by several researchers including \citep{robinson1987utility,werbos1988generalization,mozer1989focused}

However, as recurrent neural network usually has a more complex cost surface, naive backpropagation may not work well. Later in this paper, we will see that the recurrent structure introduces some critical problems, for example, the vanishing gradient problem, which makes optimization for RNN a great challenge in the society. 

\subsection{Bidirectional Recurrent Neural Network}
If we unfold an RNN, then we can get the structure of a feedforward neural network with infinite depth. Therefore, we can build a conceptual connection between RNN and feedforward network with infinite layers. Then since through the neural network history, bidirectional neural networks have been playing important roles (like Hopfield Network, RBM, DBM), a follow-up question is that what recurrent structures that correspond to the infinite layer of bidirectional models are. The answer is Bidirectional Recurrent Neural Network. 

Bidirectional Recurrent Neural Network (BRNN) was invented by \cite{schuster1997bidirectional} with the goal to introduce a structure that was unfolded to be a bidirectional neural network. Therefore, when it is applied to time series data, not only the information can be passed following the natural temporal sequences, but the further information can also reversely provide knowledge to previous time steps. 

\begin{figure}
\centering
\includegraphics[width=0.9\textwidth]{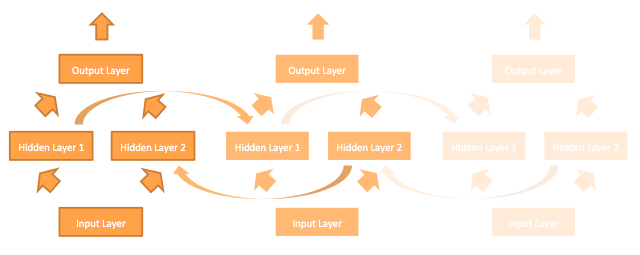}
\caption{The unfolded structured of BRNN. The temporal order is from left to right. Hidden layer 1 is unfolded in the standard way of an RNN. Hidden layer 2 is unfolded to simulate the reverse connection.}
\label{fig:brnn}
\end{figure}

Figure~\ref{fig:brnn} shows the unfolded structure of a BRNN. Hidden layer 1 is unfolded in the standard way of an RNN. Hidden layer 2 is unfolded to simulate the reverse connection. Transparency (in Figure~\ref{fig:brnn}) is applied to emphasize that unfolding an RNN is only a concept that is used for illustration purpose. The actual model handles data from different time steps with the same single model. 

BRNN is formulated as following: 
\begin{align*}
h_1^t &= \sigma(W_{h1}X+W_{r1}h_1^{t-1}) \\
h_2^t &= \sigma(W_{h2}X+W_{r2}h_2^{t+1}) \\
y &= \sigma(W_{y1}h_1^{t} + W_{y2}h_2^t)
\end{align*}
where the subscript $1$ and $2$ denote the variables associated with hidden layer $1$ and $2$ respectively. 

With the introduction of ``recurrent'' connections back from the future, Backpropagation through Time is no longer directly feasible. The solution is to treat this model as a combination of two RNNs: a standard one and a reverse one, then apply BPTT onto each of them. Weights are updated simultaneously once two gradients are computed.   

\subsection{Long Short-Term Memory}
Another breakthrough in RNN family was introduced in the same year as BRNN. \cite{hochreiter1997long} introduced a new neuron for RNN family, named Long Short-Term Memory (LSTM). When it was invented, the term ``LSTM'' is used to refer the algorithm that is designed to overcome vanishing gradient problem, with the help of a special designed \textbf{memory cell}. Nowadays, ``LSTM'' is widely used to denote any recurrent network that with that memory cell, which is nowadays referred as an LSTM cell. 

\begin{figure}
\centering
\subfloat[LSTM ``memory'' cell]{
  \includegraphics[clip,width=0.35\textwidth]{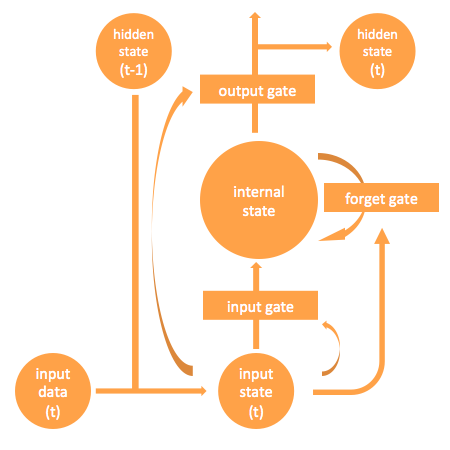}
}
\qquad
\subfloat[Input data and previous hidden state form into input state]{
  \includegraphics[clip,width=0.35\textwidth]{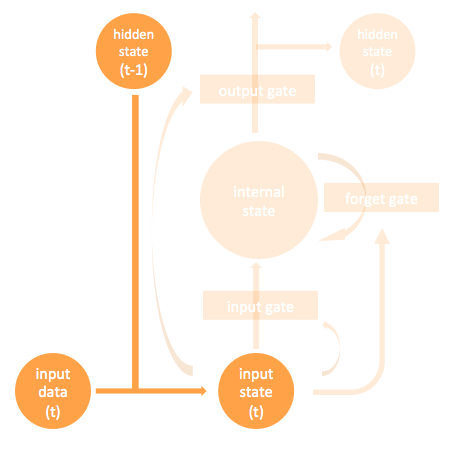}
}
\qquad
\subfloat[Calculating input gate and forget gate]{
  \includegraphics[clip,width=0.35\textwidth]{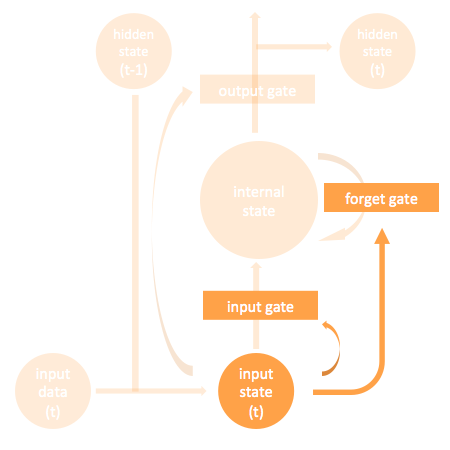}
}
\qquad
\subfloat[Calculating output gate]{
  \includegraphics[clip,width=0.35\textwidth]{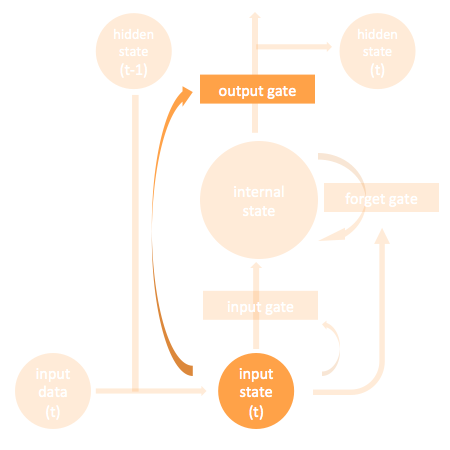}
}
\qquad
\subfloat[Update internal state]{
  \includegraphics[clip,width=0.35\textwidth]{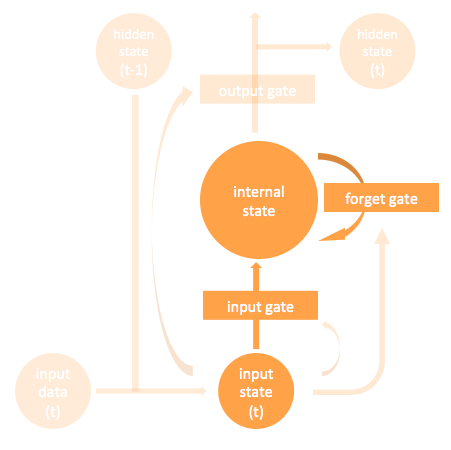}
}
\qquad
\subfloat[Output and update hidden state]{
  \includegraphics[clip,width=0.35\textwidth]{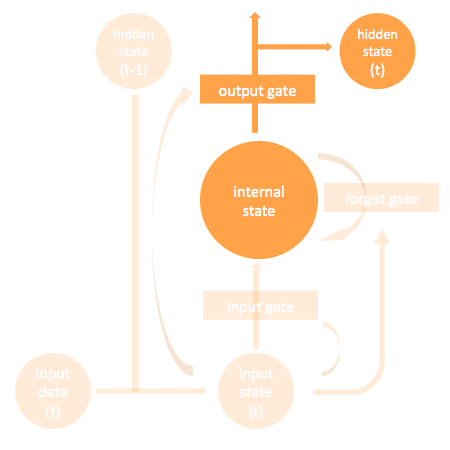}
}
\caption{The LSTM cell and its detailed functions.}
\label{fig:lstm}
\end{figure}

LSTM was introduced to overcome the problem that RNNs cannot long term dependencies \citep{bengio1994learning}. To overcome this issue, it requires the specially designed memory cell, as illustrated in Figure~\ref{fig:lstm} (a). 

LSTM consists of several critical components.  
\begin{itemize}
\item states: values that are used to offer the information for output. 
\begin{itemize}
\item [$\star$] input data: it is denoted as $x$.
\item [$\star$] hidden state: values of previous hidden layer. This is the same as traditional RNN. It is denoted as $h$.
\item [$\star$] input state: values that are (linear) combination of hidden state and input of current time step. It is denoted as $i$, and we have: 
\begin{align}
i^t = \sigma (W_{ix}x^t+W_{ih}h^{t-1})
\label{eq:lstmi}
\end{align}
\item [$\star$] internal state: Values that serve as ``memory''. It is denoted as $m$
\end{itemize}
\item gates: values that are used to decide the information flow of states. 
\begin{itemize}
\item [$\star$] input gate: it decides whether input state enters internal state. It is denoted as $g$, and we have:
\begin{align}
g^t = \sigma (W_{gi}i^t)
\label{eq:lstmg}
\end{align}
\item [$\star$] forget gate: it decides whether internal state forgets the previous internal state. It is denoted as $f$, and we have:
\begin{align}
f^t = \sigma (W_{fi}i^t)
\label{eq:lstmf}
\end{align}
\item [$\star$] output gate: it decides whether internal state passes its value to output and hidden state of next time step. It is denoted as $o$ and we have:
\begin{align}
o^t = \sigma (W_{oi}i^t)
\label{eq:lstmo}
\end{align}
\end{itemize}
\end{itemize}
Finally, considering how gates decide the information flow of states, we have the last two equations to complete the formulation of LSTM:
\begin{align}
m^t =& g^t\odot i^t + f^t m^{t-1}
\label{eq:lstmm}
\end{align}
\begin{align}
h^{t} =& o^t\odot m^t
\label{eq:lstmh}
\end{align}
where $\odot$ denotes element-wise product. 

Figure~\ref{fig:lstm} describes the details about how LSTM cell works. Figure~\ref{fig:lstm} (b) shows that how the input state is constructed, as described in Equation~\ref{eq:lstmi}. Figure~\ref{fig:lstm} (c) shows how input gate and forget gate are computed, as described in Equation~\ref{eq:lstmg} and Equation~\ref{eq:lstmf}. Figure~\ref{fig:lstm} (d) shows how output gate is computed, as described in Equation~\ref{eq:lstmo}. Figure~\ref{fig:lstm} (e) shows how internal state is updated, as described in Equation~\ref{eq:lstmm}. Figure~\ref{fig:lstm} (f) shows how output and hidden state are updated, as described in Equation~\ref{eq:lstmh}. 

All the weights are parameters that need to be learned during training. Therefore, theoretically, LSTM can learn to memorize long time dependency if necessary and can learn to forget the past when necessary, making itself a powerful model. 

With this important theoretical guarantee, many works have been attempted to improve LSTM. For example, \cite{gers2000recurrent} added a peephole connection that allows the gate to use information from the internal state. \cite{cho2014learning} introduced the Gated Recurrent Unit, known as GRU, which simplified LSTM by merging internal state and hidden state into one state, and merging forget gate and input gate into a simple update gate. Integrating LSTM cell into bidirectional RNN is also an intuitive follow-up to look into \citep{graves2013hybrid}. 

Interestingly, despite the novel LSTM variants proposed now and then, \cite{greff2015lstm} conducted a large-scale experiment investigating the performance of LSTMs and got the conclusion that none of the variants can improve upon the standard LSTM architecture significantly. Probably, the improvement of LSTM is in another direction rather than updating the structure inside a cell. Attention models seem to be a direction to go. 

\subsection{Attention Models}
Attention Models are loosely based on a bionic design to simulate the behavior of human vision attention mechanism: when humans look at an image, we do not scan it bit by bit or stare at the whole image, but we focus on some major part of it and gradually build the context after capturing the gist. Attention mechanisms were first discussed by \cite{larochelle2010learning} and \cite{denil2012learning}. The attention models mostly refer to the models that were introduced in \citep{bahdanau2014neural} for machine translation and soon applied to many different domains like \citep{chorowski2015attention} for speech recognition and \citep{xu2015show} for image caption generation. 

Attention models are mostly used for sequence output prediction. Instead of seeing the whole sequential data and make one single prediction (for example, language model), the model needs to make a sequential prediction for the sequential input for tasks like machine translation or image caption generation. Therefore, the attention model is mostly used to answer the question on where to pay attention to based on previously predicted labels or hidden states. 

The output sequence may not have to be linked one-to-one to the input sequence, and the input data may not even be a sequence. Therefore, usually, an encoder-decoder framework \citep{cho2015describing} is necessary. The encoder is used to encode the data into representations and decoder is used to make sequential predictions. Attention mechanism is used to locate a region of the representation for predicting the label in current time step. 

\begin{figure}
\centering
\includegraphics[width=0.8\textwidth]{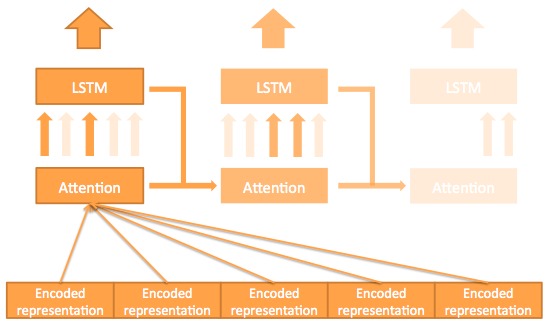}
\caption{The unfolded structured of an attention model. Transparency is used to show that unfolding is only conceptual. The representation encoder learns are all available to the decoder across all time steps. Attention module only selects some to pass onto LSTM cell for prediction. }
\label{fig:attention}
\end{figure}

Figure~\ref{fig:attention} shows a basic attention model under encoder-decoder network structure. The representation encoder encodes is all accessible to attention model, and attention model only selects some regions to pass onto the LSTM cell for further usage of prediction making. 

Therefore, all the magic of attention models is about how this attention module in Figure~\ref{fig:attention} helps to localize the informative representations. 

To formalize how it works, we use $r$ to denote the encoded representation (there is a total of $M$ regions of representation), use $h$ to denote hidden states of LSTM cell. Then, the attention module can generate the unscaled weights for $i$th region of the encoded representation as:
\begin{align*}
\beta_i^t = f(h^{t-1}, r, \{\alpha^{t-1}_j\}_{j=1}^M)
\end{align*}
where $\alpha^{t-1}_j$ is the attention weights computed at the
previous time step, and can be computed at current time step as a simple softmax function:
\begin{align*}
\alpha_i^t = \dfrac{\exp(\beta_i^t)}{\sum_j^M\exp(\beta_j^t)}
\end{align*}
Therefore, we can further use the weights $\alpha$ to reweight the representation $r$ for prediction. There are two ways for the representation to be reweighted:
\begin{itemize}
\item Soft attention: The result is a simple weighted sum of the context vectors such that:
\begin{align*}
r^t = \sum_j^M\alpha_j^tc_j
\end{align*}
\item Hard attention: The model is forced to make a hard decision by only localizing one region: sampling one region out following multinoulli distribution. 
\end{itemize}

One problem about hard attention is that sampling out of multinoulli distribution is not differentiable. Therefore, the gradient based method can be hardly applied. Variational methods \citep{ba2014multiple} or policy gradient based method \citep{sutton1999policy} can be considered.  

\subsection{Deep RNNs and the future of RNNs}
In this very last section of the evolutionary path of RNN family, we will visit some ideas that have not been fully explored. 

\subsubsection{Deep Recurrent Neural Network}
Although recurrent neural network suffers many issues that deep neural network has because of the recurrent connections, current RNNs are still not deep models regarding representation learning compared to models in other families. 

\begin{figure}
\centering
\subfloat[Deep input architecture]{
  \includegraphics[clip,width=0.25\textwidth]{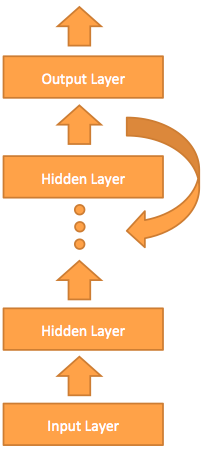}
}
\qquad
\subfloat[Deep recurrent architecture]{
  \includegraphics[clip,width=0.25\textwidth]{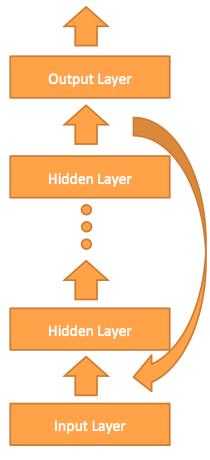}
}
\qquad
\subfloat[Deep output architecture]{
  \includegraphics[clip,width=0.25\textwidth]{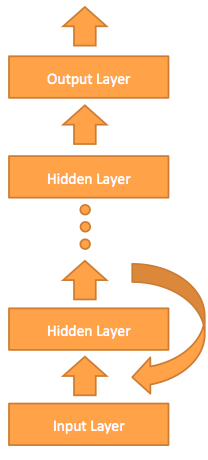}
}
\caption{Three different formulations of deep recurrent neural network. }
\label{fig:drnn}
\end{figure}

\cite{pascanu2013construct} formalizes the idea of constructing deep RNNs by extending current RNNs. Figure~\ref{fig:drnn} shows three different directions to construct a deep recurrent neural network by increasing the layers of the input component (Figure~\ref{fig:drnn} (a)), recurrent component (Figure~\ref{fig:drnn} (b)) and output component (Figure~\ref{fig:drnn} (c)) respectively. 

\subsubsection{The Future of RNNs}
RNNs have been improved in a variety of different ways, like assembling the pieces together with Conditional Random Field \citep{yang2016multi}, and together with CNN components \citep{ma2016end}. In addition, convolutional operation can be directly built into LSTM, resulting ConvLSTM \citep{xingjian2015convolutional}, and then this ConvLSTM can be also connected with a variety of different components \citep{de2016dynamic,kalchbrenner2016video}.  

One of the most fundamental problems of training RNNs is the vanishing/exploding gradient problem, introduced in detail in \citep{bengio1994learning}. The problem basically states that for traditional activation functions, the gradient is bounded. When gradients are computed by backpropagation following chain rule, the error signal decreases exponentially within the time steps the BPTT can trace back, so the long-term dependency is lost. LSTM and ReLU are known to be good solutions for vanishing/exploding gradient problem. However, these solutions introduce ways to bypass this problem with clever design, instead of solving it fundamentally. While these methods work well practically, the fundamental problem for a general RNN is still to be solved. \cite{pascanu2013difficulty} attempted some solutions, but there are still more to be done.

\newpage
\section{Optimization of Neural Networks}
\label{sec:opt}

The primary focus of this paper is deep learning models. However, optimization is an inevitable topic in the development history of deep learning models. In this section, we will briefly revisit the major topics of optimization of neural networks. During our introduction of the models, some algorithms have been discussed along with the models. Here, we will only discuss the remaining methods that have not been mentioned previously. 

\subsection{Gradient Methods}
\label{sec:grad}
Despite the fact that neural networks have been developed for over fifty years, the optimization of neural networks still heavily rely on gradient descent methods within the algorithm of backpropagation. This paper does not intend to introduce the classical backpropagation, gradient descent method and its stochastic version and batch version and simple techniques like momentum method, but starts right after these topics. 

Therefore, the discussion of following gradient methods starts from the vanilla gradient descent as following:
\begin{align*}
\theta^{t+1} = \theta^t -\eta \bigtriangledown_\theta^t
\end{align*}
where $\bigtriangledown_\theta$ is the gradient of the parameter $\theta$, $\eta$ is a hyperparameter, usually known as learning rate. 

\subsubsection{Rprop}
Rprop was introduced by \cite{riedmiller1993direct}. It is a unique method even studied back today as it does not fully utilize the information of gradient, but only considers the sign of it. In other words, it updates the parameters following:
\begin{align*}
\theta^{t+1} = \theta^t -\eta I(\bigtriangledown_\theta^t>0) + \eta I(\bigtriangledown_\theta^t<0)
\end{align*}
where $I(\cdot)$ stands for an indicator function. 

This unique formalization allows the gradient method to overcome some cost curvatures that may not be easily solved with today's dominant methods. This two-decade-old method may be worth some further study these days. 

\subsubsection{AdaGrad}
AdaGrad was introduced by \cite{duchi2011adaptive}. It follows the idea of introducing an adaptive learning rate mechanism that assigns higher learning rate to the parameters that have been updated more mildly and assigns lower learning rate to the parameters that have been updated dramatically. The measure of the degree of the update applied is the $\ell_{2}$ norm of historical gradients, $S^t=||\bigtriangledown_\theta^1, \bigtriangledown_\theta^2, ... \bigtriangledown_\theta^t||_2^2$, therefore we have the update rule as following:
\begin{align*}
\theta^{t+1} = \theta^t -\dfrac{\eta}{S^t+\epsilon} \bigtriangledown_\theta^t
\end{align*}
where $\epsilon$ is small term to avoid $\eta$ divided by zero. 

AdaGrad has been showed with great improvement of robustness upon traditional gradient method \citep{dean2012large}. However, the problem is that as $\ell_{2}$ norm accumulates, the fraction of $\eta$ over $\ell_{2}$ norm decays to a substantial small term. 

\subsubsection{AdaDelta}
AdaDelta is an extension of AdaGrad that aims to reducing the decaying rate of learning rate, proposed in \citep{zeiler2012adadelta}. Instead of accumulating the gradients of each time step as in AdaGrad, AdaDelta re-weights previously accumulation before adding current term onto previously accumulated result, resulting in:
\begin{align*}
(S^t)^2 = \beta(S^{t-1})^2+(1-\beta)(\bigtriangledown_\theta^t)^2
\end{align*}
where $\beta$ is the weight for re-weighting. Then the update rule is the same as AdaGrad:
\begin{align*}
\theta^{t+1} = \theta^t -\dfrac{\eta}{S^t+\epsilon} \bigtriangledown_\theta^t
\end{align*}
which is almost the same as another famous gradient variant named RMSprop\footnote{It seems this method  never gets published, the resources all trace back to Hinton's slides at http://www.cs.toronto.edu/\~tijmen/csc321/slides/lecture\_slides\_lec6.pdf}. 

\subsubsection{Adam}
Adam stands for Adaptive Moment Estimation, proposed in \citep{kingma2014adam}. Adam is like a combination momentum method and AdaGrad method, but each component are re-weighted at time step $t$. Formally, at time step $t$, we have:
\begin{align*}
\Delta_\theta^t =& \alpha\Delta_\theta^{t-1}+(1-\alpha)\bigtriangledown_\theta^t\\
(S^t)^2 =& \beta(S^{t-1})^2+(1-\beta)(\bigtriangledown_\theta^t)^2\\
\theta^{t+1} =& \theta^t - \dfrac{\eta}{S^t+\epsilon}\Delta_\theta^t
\end{align*}

All these modern gradient variants have been published with a promising claim that is helpful to improve the convergence rate of previous methods. Empirically, these methods seem to be indeed helpful, however, in many cases, a good choice of these methods seems only to benefit to a limited extent. 

\subsection{Dropout}
Dropout was introduced in \citep{hinton2012improving,srivastava2014dropout}. The technique soon got influential, not only because of its good performance but also because of its simplicity of implementation. The idea is very simple: randomly dropping out some of the units while training. More formally: on each training case, each hidden unit is randomly omitted from the network with a probability of $p$.

As suggested by \cite{hinton2012improving}, Dropout can be seen as an efficient way to perform model averaging across a large number of different neural networks, where overfitting can be avoided with much less cost of computation. 

Because of the actual performance it introduces, Dropout soon became very popular upon its introduction, a lot of work has attempted to understand its mechanism in different perspectives, including \citep{baldi2013understanding,cho2013understanding,ma2016dropout}. It has also been applied to train other models, like SVM \citep{chen2014dropout}. 

\subsection{Batch Normalization and Layer Normalization}
Batch Normalization, introduced by \cite{ioffe2015batch}, is another breakthrough of optimization of deep neural networks. They addressed the problem they named as \textit{internal covariate shift}. Intuitively, the problem can be understood as the following two steps: 1) a learned function is barely useful if its input changes (In statistics, the input of a function is sometimes denoted as covariates). 2) each layer is a function and the changes of parameters of below layers change the input of current layer. This change could be dramatic as it may shift the distribution of inputs. 

\cite{ioffe2015batch} proposed the Batch Normalization to solve this issue, formally following the steps:
\begin{align*}
\mu_B =& \dfrac{1}{n}\sum_{i=1}^nx_i \\
\sigma_B^2 =& \dfrac{1}{n}\sum_{i=1}^n(x_i-\mu_B)^2\\
\hat{x_i} =& \dfrac{x_i-\mu_B}{\sigma_B+\epsilon}\\
y_i =& \sigma_L\hat{x_i} + \mu_L \\
\end{align*}
where $\mu_B$ and $\sigma_B$ denote the mean and variance of that batch. $\mu_L$ and $\sigma_L$ two parameters learned by the algorithm to rescale and shift the output. $x_i$ and $y_i$ are inputs and outputs of that function respectively. 

These steps are performed for every batch during training. Batch Normalization turned out to work very well in training empirically and soon became popularly. 

As a follow-up, \cite{ba2016layer} proposes the technique Layer Normalization, where they ``transpose'' batch normalization into layer normalization by computing the mean and variance used for normalization from all of the summed inputs to the neurons in a layer on a single training case. Therefore, this technique has a nature advantage of being applicable to recurrent neural network straightforwardly. However, it seems that this ``transposed batch normalization'' cannot be implemented as simple as Batch Normalization. Therefore, it has not become as influential as Batch Normalization is. 

\subsection{Optimization for ``Optimal'' Model Architecture}
In the very last section of optimization techniques for neural networks, we revisit some old methods that have been attempted with the aim to learn the ``optimal'' model architecture. Many of these methods are known as constructive network approaches. Most of these methods have been proposed decades ago and did not raise enough impact back then. Nowadays, with more powerful computation resources, people start to consider these methods again. 

Two remarks need to be made before we proceed: 1) Obviously, most of these methods can trace back to counterparts in non-parametric machine learning field, but because most of these methods did not perform enough to raise an impact, focusing a discussion on the evolutionary path may mislead readers. Instead, we will only list these methods for the readers who seek for inspiration. 2) Many of these methods are not exclusively optimization techniques because these methods are usually proposed with a particularly designed architecture. Technically speaking, these methods should be distributed to previous sections according to the models associated. However, because these methods can barely inspire modern modeling research, but may have a chance to inspire modern optimization research, we list these methods in this section. 

\subsubsection{Cascade-Correlation Learning}
One of the earliest and most important works on this topic was proposed by \cite{fahlman1989cascade}. They introduced a model, as well as its corresponding algorithm named Cascade-Correlation Learning. The idea is that the algorithm starts with a minimum network and builds up towards a bigger network. Whenever another hidden unit is added, the parameters of previous hidden units are fixed, and the algorithm only searches for an optimal parameter for the newly-added hidden unit. 

Interestingly, the unique architecture of Cascade-Correlation Learning grants the network to grow deeper and wider at the same time because every newly added hidden unit takes the data together with outputs of previously added units as input. 

Two important questions of this algorithm are 1) when to fix the parameters of current hidden units and proceed to add and tune a newly added one 2) when to terminate the entire algorithm. These two questions are answered in a similar manner: the algorithm adds a new hidden unit when there are no significant changes in existing architecture and terminates when the overall performance is satisfying. This training process may introduce problems of overfitting, which might account for the fact that this method is seen much in modern deep learning research. 

\subsubsection{Tiling Algorithm}
\cite{mezard1989learning} presented the idea of Tiling Algorithm, which learns the parameters, the number of layers, as well as the number of hidden units in each layer simultaneously for feedforward neural network on Boolean functions. Later this algorithm was extended to multiple class version by \cite{parekh1997constructive}. 

The algorithm works in such a way that on every layer, it tries to build a layer of hidden units that can cluster the data into different clusters where there is only one label in one cluster. The algorithm keeps increasing the number of hidden units until such a clustering pattern can be achieved and proceed to add another layer. 

\cite{mezard1989learning} also offered a proof of theoretical guarantees for Tiling Algorithm. Basically, the theorem says that Tiling Algorithm can greedily improve the performance of a neural network. 

\subsubsection{Upstart Algorithm}
\cite{frean1990upstart} proposed the Upstart Algorithm. Long story short, this algorithm is simply a neural network version of the standard decision tree \citep{safavian1990survey} where each tree node is replaced with a linear perceptron. Therefore, the tree is seen as a neural network because it uses the core component of neural networks as a tree node. As a result, standard way of building a tree is advertised as building a neural network automatically. 

Similarly, \cite{bengio2005convex} proposed a boosting algorithm where they replace the weak classifier as neurons. 

\subsubsection{Evolutionary Algorithm}
Evolutionary Algorithm is a family of algorithms uses mechanisms inspired by biological evolution to search in a parameter space for the optimal solution. Some prominent examples in this family are genetic algorithm \citep{mitchell1998introduction}, which simulates natural selection and ant colony optimization algorithm \citep{colorni1991distributed}, which simulates the cooperation of an ant colony to explore surroundings. 

\cite{yao1999evolving} offered an extensive survey of the usage of evolution algorithm upon the optimization of neural networks, in which Yao introduced several encoding schemes that can enable the neural network architecture to be learned with evolutionary algorithms. The encoding schemes basically transfer the network architecture into vectors, so that a standard algorithm can take it as input and optimize it. 

So far, we discussed some representative algorithms that are aimed to learn the network architecture automatically. Most of these algorithms eventually fade out of modern deep learning research, we conjecture two main reasons for this outcome: 1) Most of these algorithms tend to overfit the data. 2) Most of these algorithms are following a greedy search paradigm, which will be unlikely to find the optimal architecture. 

However, with the rapid development of machine learning methods and computation resources in the last decade, we hope these constructive network methods we listed here can still inspire the readers for substantial contributions to modern deep learning research.

\newpage

\section{Conclusion}
In this paper, we have revisited the evolutionary path of the nowadays deep learning models. We revisited the paths for three major families of deep learning models: the deep generative model family, convolutional neural network family, and recurrent neural network family as well as some topics for optimization techniques.

This paper could serve two goals: 1) First, it documents the major milestones in the science history that have impacted the current development of deep learning. These milestones are not limited to the development in computer science fields. 2) More importantly, by revisiting the evolutionary path of the major milestone, this paper should be able to suggest the readers that how these remarkable works are developed among thousands of other contemporaneous publications. Here we briefly summarize three directions that many of these milestones pursue:
\begin{itemize}
\item \textbf{Occam's razor:} While it seems that part of the society tends to favor more complex models by layering up one architecture onto another and hoping backpropagation can find the optimal parameters, history says that masterminds tend to think simple: Dropout is widely recognized not only because of its performance, but more because of its simplicity in implementation and intuitive (tentative) reasoning. From Hopfield Network to Restricted Boltzmann Machine, models are simplified along the iterations until when RBM is ready to be piled-up. 
\item \textbf{Be ambitious:} If a model is proposed with substantially more parameters than contemporaneous ones, it must solve a problem that no others can solve nicely to be remarkable. LSTM is much more complex than traditional RNN, but it bypasses the vanishing gradient problem nicely. Deep Belief Network is famous not due to the fact the they are the first one to come up with the idea of putting one RBM onto another, but due to that they come up an algorithm that allow deep architectures to be trained effectively. 
\item \textbf{Widely read:} Many models are inspired by domain knowledge outside of machine learning or statistics field. Human visual cortex has greatly inspired the development of convolutional neural networks. Even the recent popular Residual Networks can find corresponding mechanism in human visual cortex. Generative Adversarial Network can also find some connection with game theory, which was developed fifty years ago. 
\end{itemize}

We hope these directions can help some readers to impact more on current society. More directions should also be able to be summarized through our revisit of these milestones by readers.  

\section*{Acknowledgements}
Thanks to the demo from http://beej.us/blog/data/convolution-image-processing/ for a quick generation of examples in Figure~\ref{fig:kernel}. 
Thanks to Bojian Han at Carnegie Mellon University for the examples in Figure~\ref{fig:imagenet}. 
Thanks to the blog at http://sebastianruder.com/optimizing-gradient-descent/index.html for a summary of gradient methods in Section~\ref{sec:grad}.
Thanks to Yutong Zheng and Xupeng Tong at Carnegie Mellon University for suggesting some relevant contents. 

\vskip 0.2in
\bibliography{ref}

\end{document}